\documentclass{article}

\usepackage{microtype}
\usepackage{graphicx}
\usepackage{subfigure}
\usepackage{rotating}

\usepackage{booktabs} 

\usepackage{hyperref}

\usepackage{multirow}

\usepackage[accepted]{icml2023}

\usepackage{amsmath}
\usepackage{amssymb}
\usepackage{mathtools}
\usepackage{amsthm}
\usepackage{xspace}
\usepackage[capitalize,noabbrev]{cleveref}

\usepackage{times}
\usepackage{latexsym}
\usepackage{bbm}

\usepackage[T1]{fontenc}
\usepackage[utf8]{inputenc}

\theoremstyle{plain}

\theoremstyle{definition}

\theoremstyle{remark}

\newif\ifhidecomments

\ifhidecomments
\usepackage{environ}
\NewEnviron{hide}{}

\fi

\newcommand\cbtm{\textsc{c-BTM}\xspace}

\newcommand\elm{\textsc{ELM}\xspace}

\usepackage[textsize=tiny]{todonotes}

\icmltitlerunning{Scaling Expert Language Models with Unsupervised Domain Discovery}
\begin{document}

\twocolumn[
\icmltitle{Scaling Expert Language Models with Unsupervised Domain Discovery}



\icmlsetsymbol{equal}{*}
\icmlsetsymbol{wmeta}{$\dagger$}

\begin{icmlauthorlist}
\icmlauthor{Suchin Gururangan}{equal,uw,wmeta}
\icmlauthor{Margaret Li}{equal,uw,meta} 
\icmlauthor{Mike Lewis}{meta} 
\icmlauthor{Weijia Shi}{uw,meta} 
\icmlauthor{Tim Althoff}{uw} \\ \icmlauthor{Noah A. Smith}{uw,allenai}
\icmlauthor{Luke Zettlemoyer}{uw,meta}

\end{icmlauthorlist}

\icmlaffiliation{uw}{University of Washington}
\icmlaffiliation{meta}{Meta AI}
\icmlaffiliation{allenai}{Allen Institute for Artificial Intelligence}

\icmlcorrespondingauthor{Suchin Gururangan}{sg01@cs.washington.edu}
\icmlcorrespondingauthor{Margaret Li}{margsli@cs.washington.edu}

\icmlkeywords{Machine Learning, ICML}

\vskip 0.3in
]




\printAffiliationsAndNotice{\icmlEqualContribution} 

\begin{abstract}



Large language models are typically trained densely: all parameters are updated with respect to all inputs. This requires synchronization of billions of parameters across thousands of GPUs.
We introduce a simple but effective method to \emph{asynchronously} train large, sparse language models on arbitrary text corpora.
Our method clusters a corpus into sets of related documents, trains a separate expert language model on each cluster, and combines them in a sparse ensemble for inference.
This approach generalizes embarrassingly parallel training by automatically discovering the domains for each expert, and eliminates nearly all the communication overhead of existing sparse language models.
Our technique outperforms dense baselines on multiple corpora and few-shot tasks, and our analysis shows that specializing experts to meaningful clusters is key to these gains.
Performance also improves with the number of experts and size of training data, suggesting this is a highly efficient and accessible approach to training large language models. 
\end{abstract}

\section{Introduction}
\label{sec:introduction}

\begin{figure}[t!]
    \centering
    \includegraphics[width=\columnwidth]{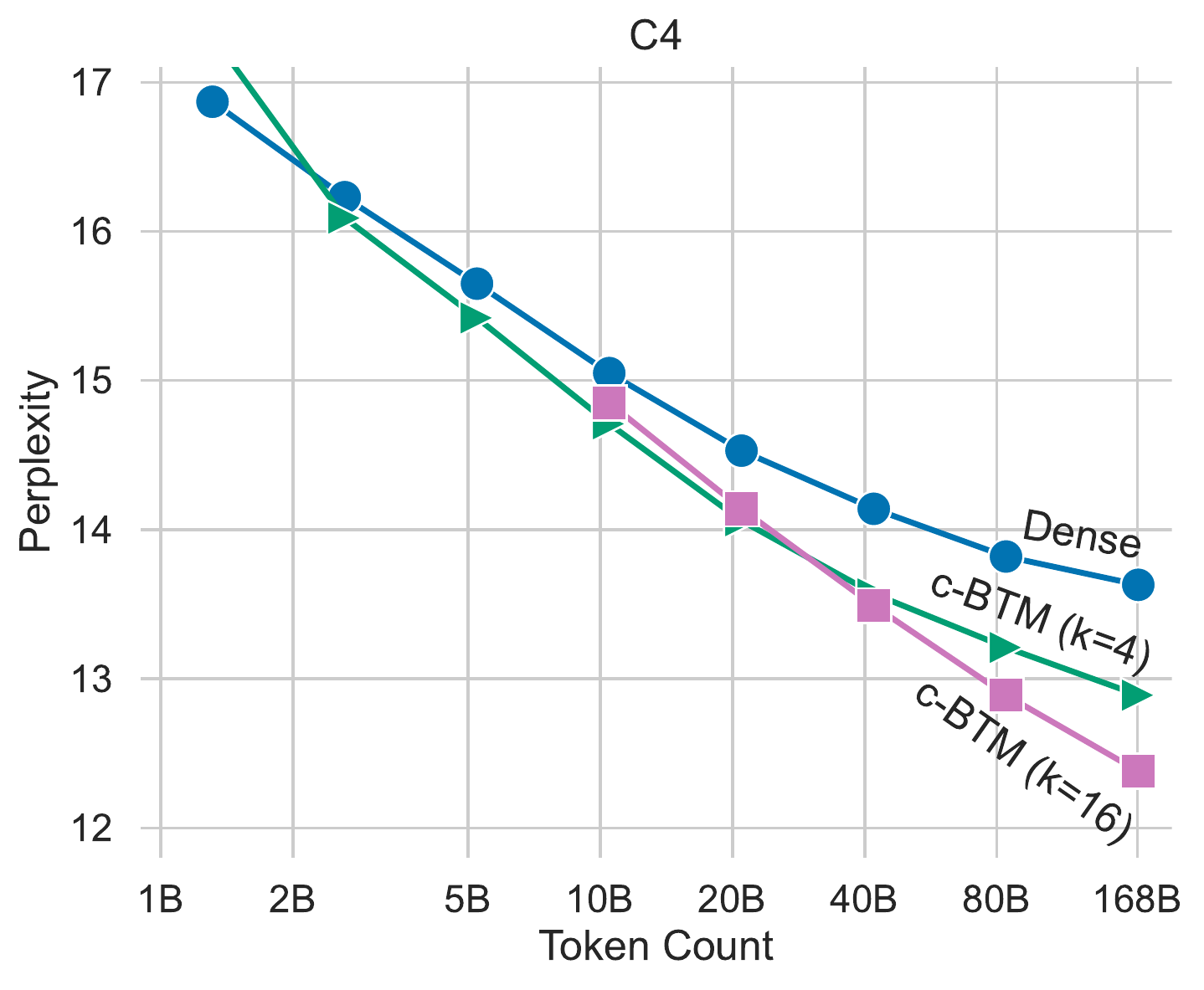} 
    \vspace{-2em}
    \caption{\textbf{We present \cbtm, a new technique to asynchronously scale expert LMs (\S\ref{sec:cbtm}).} \cbtm splits a corpus into $k$ clusters, trains an expert LM on each cluster, and creates a sparse ensemble during inference. Above, LMs trained with \cbtm (with 4 or 16 clusters) achieve lower validation perplexity than compute-matched dense LMs. These LMs begin with OPT-1.3B \citep{opt}, and are further trained on C4 \citep{t5}. The optimal cluster count for \cbtm, and its performance gains, increase with the size of training data (shown in log-scale).} 
    \label{fig:schwartzplot}
\end{figure}

Language models (LMs) are trained on up to trillions of tokens of text \citep{hoffmann2022training, llama}. This improves performance on many tasks, but also incurs an extreme cost: thousands of GPUs need to be active simultaneously to update all parameters at each step \citep{opt, palm}.  
Branch-Train-Merge (BTM; \citealt{btm}) alleviates this cost by dividing the total compute among a collection of smaller expert language models (ELMs), each independently trained on a distinct subset (or \emph{domain}) of the training corpus and ensembled during inference. However, BTM relies on document metadata to identify domains, and such supervision is not always available \citep[e.g., in large Internet crawls;][]{t5, gopher, pile}.  Moreover, the optimal number of metadata-based domains for a fixed budget is unknown, since metadata cannot be easily merged or divided. 




In this work, we introduce Cluster-Branch-Train-Merge (\cbtm; Figure \ref{fig:schwartzplot}), a metadata-free algorithm to scale LMs without massive multi-node synchronization. We use unsupervised clustering to discover domains in a corpus, and  train an \elm{} on each cluster independently (\S\ref{sec:cbtm_iteration}).
At inference time, we \emph{sparsely} activate a subset of the trained \elm{}s (\S\ref{sec:cbtm_inference}). We ensemble \elm{}s by weighting their outputs with the distances between an embedding of the current context and each expert's cluster center. This enables simple and efficient sparse computation \citep{https://doi.org/10.48550/arxiv.2209.01667} by retrieving only the top-$k$ experts when predicting each new token. 

\begin{figure*}[t!]
    \centering
    \includegraphics[width=\textwidth]{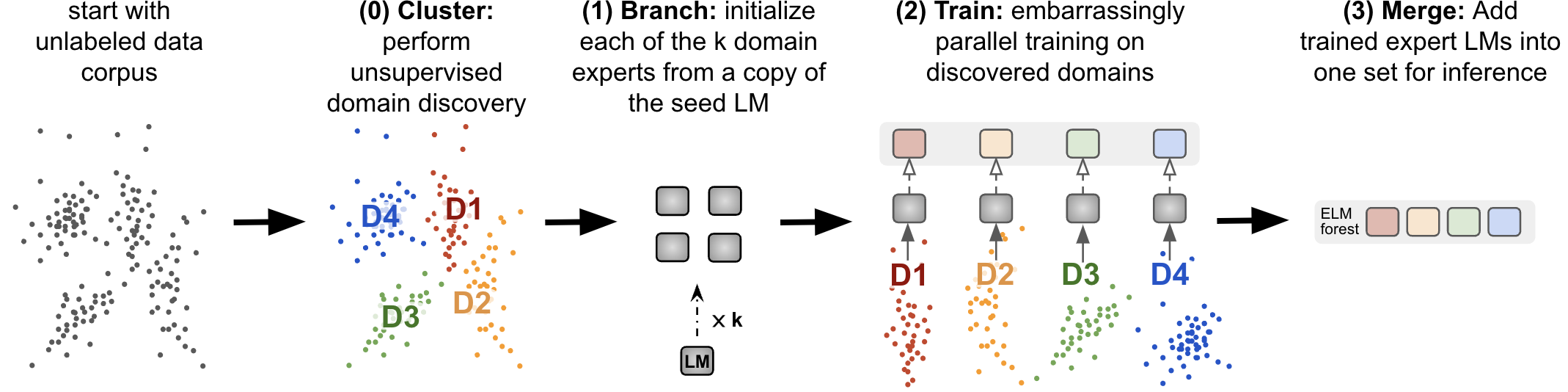} \vspace{-2em}
    \caption{
    \textbf{\cbtm training process (\S\ref{sec:cbtm_iteration}).} \cbtm begins with unsupervised domain discovery using $k$-means clustering. We then initialize expert language models (ELMs) with a seed language model (e.g., OPT; \citealt{opt}) and train an ELM on each cluster. The resulting experts are added to a larger collection for sparse inference.}
    \label{fig:process}
\end{figure*}


\cbtm generalizes BTM by allowing for fine-grained control over the number and size of data clusters, since they are automatically learned without being constrained by available metadata. 
We use this new capability to investigate the scaling properties of \cbtm as a function of the number of experts trained, controlling for a variety of factors (\S\ref{sec:experiments}). Extensive experiments show that training more clusters always results in better validation perplexity than single cluster (i.e., dense)  models, and the optimal cluster count increases with the overall compute (\S\ref{sec:core_results}). These results are consistent for both 1.3B and 6.7B parameter experts. 

With more clusters, we can aggressively parallelize expert training: for example, we train 128 \elm{}s (168B parameters in total) on 168B tokens of text in aggregate with only 8 GPUs at a time. This enables us to avoid many practical difficulties associated with training large LMs across many nodes simultaneously (\S\ref{sec:time_comparison}). Moreover, the number of  parameters at inference time can be kept constant even as the number of experts grows (\S\ref{sec:sparsity_analysis}): using just the top-2 or top-4 experts is comparable to using all experts, while using just the top-1 expert still outperforms the dense model. Training with more clusters is also more effective than training larger dense models: in \S\ref{sec:zflops}, we demonstrate that training many 1.3B expert LMs, and sparsifying them to a 5.2B parameter LM, achieves the same perplexity as a 6.7B dense model, but with only 29\% as many training FLOPs. These gains are also reflected in few-shot text classification experiments (\S\ref{sec:downstream_tasks}), which show that \cbtm  models outperform dense baselines even with heavily sparsified inference.

\cbtm provides a radically simplified sparse modeling approach that eliminates nearly all communication overhead from existing sparse LM schemes. Existing sparse LMs typically route different tokens to specialist parameters \cite{lepikhin2020gshard,fedus2021switch,https://doi.org/10.48550/arxiv.2202.01169}. However, they have yet to be widely adopted, perhaps due in part to the communication costs of routing each token in each sparse layer \citep{artetxe2021efficient}, challenges in learning to specialize experts to tokens \citep{https://doi.org/10.48550/arxiv.2202.09368}, and the necessity of additional mechanisms to balance expert utilization \citep{baselayers}. \cbtm improves over sparse LMs by routing sequences (instead of tokens) using offline balanced clustering (instead of online load balancing) with no shared parameters between experts. We compare directly to a mixture-of-experts model with top-2 routing \citep{lepikhin2020gshard} in \S\ref{sec:moe_comparison}.

Our final analysis (\S\ref{sec:analysis}) shows that balanced clustering is key to \cbtm performance; it works as well as expert assignment with gold metadata, and strongly outperforms random and unbalanced clustering baselines. Overall, our findings suggest that \cbtm is an efficient and accessible method to scale large language models into massive datasets. We release our code and models publicly.\footnote{\url{https://github.com/kernelmachine/cbtm}}

\section{\cbtm}
\label{sec:cbtm}

We introduce \cbtm, a method for embarrassingly parallel training that specializes expert language models to domains discovered through clustering instead of metadata. \cbtm enables scaling to arbitrary numbers of domains and compute budgets on any corpus. In this section, we outline \cbtm training (Figure \ref{fig:process}) and inference (Figure \ref{fig:inference_process}).

\subsection{Training}
\label{sec:cbtm_iteration}

\paragraph{Step 0: Cluster}

To segment our corpus, we employ $k$-means clustering, enforcing balanced clusters. ELMs trained without this constraint perform worse  (\S\ref{sec:analysis_balancing}).\footnote{Other techniques to improve clusters, e.g. $k$-means++ \citep{10.5555/1283383.1283494}, can be used to improve performance.}

Consider the iterative, hard expectation-maximization view of $k$-means clustering. In the expectation step, each document embedding is assigned to a cluster center based on its Euclidean distance to each center. In the maximization step, each cluster center is updated to be the mean embedding of the current set of documents assigned to it. To balance the clusters, we formulate the expectation step as a balanced linear assignment problem \citep{Malinen2014BalancedKF}. Given $D$ document embeddings with representations $\{w_1, \ldots, w_D\}$ and $K$ cluster centers with representations $\{h_1,\ldots,h_K\}$, we assign each document $d$ to a cluster with the assignment index $a_d \in \{0,\ldots,K\}$:
\begin{align}
\max_{a_1, \ldots, a_D} \sum_{d=1}^D -dist(h_{a_d}, w_d)
 \text{ s.t. $\forall k$, }  \sum_{d=1}^{D}\mathbbm{1}_{a_d = k} = \frac{D}{K} 
\end{align}
where $dist$ is the Euclidean distance. Many algorithms exist to solve this problem; we follow \citet{baselayers} and use the auction algorithm \citep{Bertsekas1992AuctionAF}.  We only use balancing when estimating the cluster centers; we use greedy inference when predicting clusters, as balancing at inference time is cumbersome for massive corpora. 


In our experiments, we use a simple tf-idf embedding function, which is highly efficient at scale and leads to interpretable clusters.\footnote{In initial experiments, tf-idf outperformed other scalable text embeddings, like hash embeddings \citep{https://doi.org/10.48550/arxiv.1709.03933}.}  We only use a single shard of each corpus to train our clustering model. Any new document, once embedded, can be efficiently mapped to its nearest cluster(s) without additional training. Any embedding function can be used, though the choice of embedder may apply different assumptions of what constitutes a textual domain and come with efficiency trade-offs.\footnote{tf-idf assumes that domains are lexically-driven, which may not correspond with other notions of domain.}  Comparing to other embedding or clustering methods is an interesting area for future work, and could likely improve performance. 



\paragraph{Step 1: Branch (from seed LM)} To begin training experts on each of the $k$ clusters from Step 0, we first \emph{branch} from (i.e., make $k$ copies of) a \emph{seed} LM. Seed LMs are critical for the overall functionality of ELMs, and ELMs perform best when the seed LM has been trained with a diverse corpus \citep{btm}. In our experiments, we use an OPT LM \citep{opt} as our seed.\footnote{\citealt{btm} find that dedicating more compute to branching (rather than seed training) leads to better in-domain performance, and the choice of seed LM has a strong effect on the modularity of the resulting ELMs. Future work may explore the effect of different seed LMs on \cbtm performance.}

\paragraph{Step 2: Train} We assign each \elm to a single cluster, and train on each cluster with the log likelihood objective.

\paragraph{Step 3: Merge} After training on the assigned domain, we add the new \elm into a larger collection for inference.

In this work, we focus on a single iteration of \cbtm for simplicity. Future work may explore branching from already trained experts in multiple iterations.

\subsection{Inference}
\label{sec:cbtm_inference}

\begin{figure}[t!]
    \centering
    \includegraphics[width=\columnwidth]{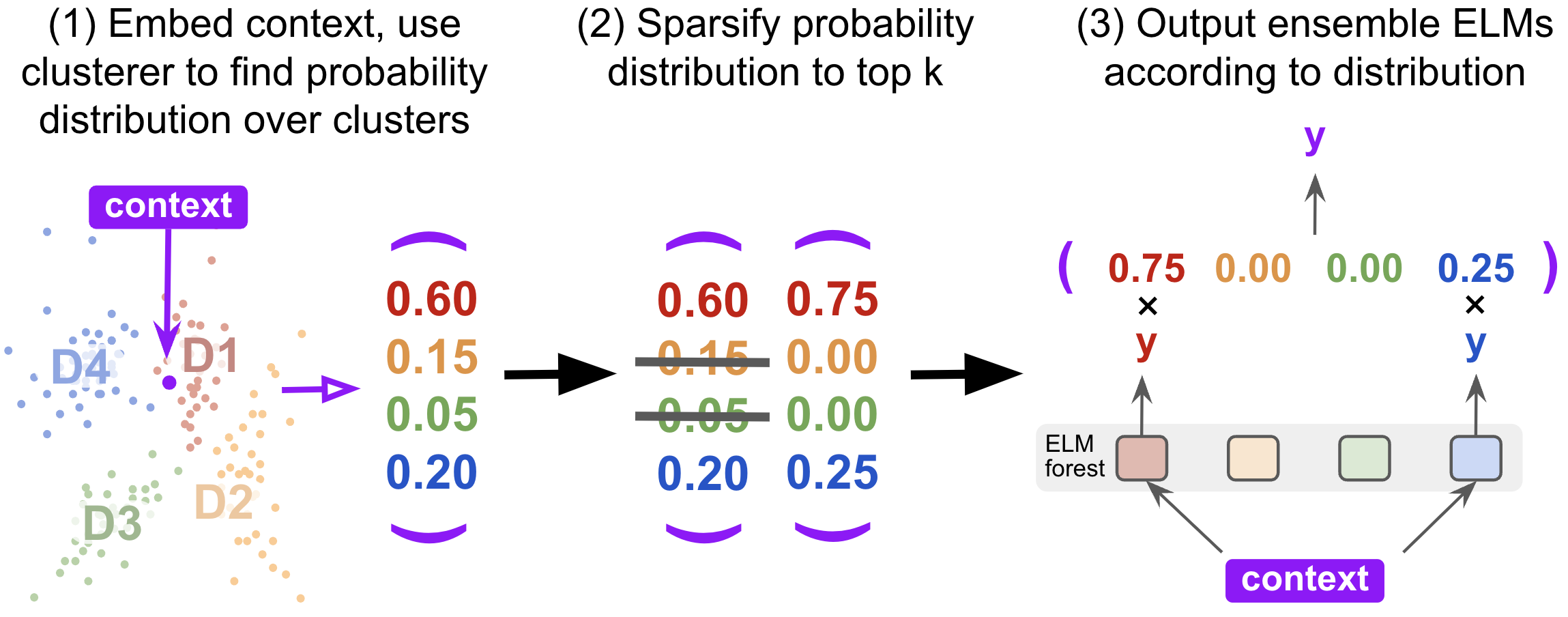}
    \vspace{-.5em}
    \caption{\textbf{\cbtm inference process (\S\ref{sec:cbtm_inference}).}  At inference time, we embed each incoming context and estimate a probability distribution over clusters, by calculating the distance between the embedded context and each cluster center. We use this probability distribution, optionally sparsified to use only the top-k experts, to weight an output ensemble of the \elm{}s.}
    \label{fig:inference_process}
\end{figure}

At inference time, we use a sparse ensemble of the outputs of ELMs for incoming test contexts (Figure \ref{fig:inference_process}).
Formally, consider that the language model provides, at each timestep, $p(X_t \mid \boldsymbol{x}_{<t})$.  We introduce a domain variable $D$, alongside each sequence. Then the next-step conditional distribution on the history $\boldsymbol{x}_{<t}$ is:
\begin{align}\small
    p(X_t \mid \boldsymbol{x}_{<t}) & \small{= \sum_{j=1}^k p(X_t \mid \boldsymbol{x}_{<t}, D = j) \cdot \underbrace{p(D = j \mid \boldsymbol{x}_{<t})}_{\mbox{ensemble weights}}}
\end{align}



With the pretrained embedder and clustering model from Step 0 (\S\ref{sec:cbtm_iteration}), we  embed the context $h_{x_{<t}}$ and use the $k$ cluster centers $\{h_{c_0} \ldots h_{c_k}\}$. We set ensemble weights as:
\begin{align}\label{eq:topk}
 p(D = j \mid \boldsymbol{x}_{<t}) \propto  \text{topk} [ \exp (-dist(h_{x_{<t}}, h_{c{_j}})^{2}/T) ]
\end{align}

Where $dist$ is the Euclidean distance, $T$ is a temperature parameter which sharpens or smoothes the probability distribution over cluster centers, and the top-k function filters for the top-$k$ probabilities and renormalizes the distribution to sum to 1. This formulation is reminiscent of nearest-neighbor retrieval mechanisms for language models  \citep{https://doi.org/10.48550/arxiv.1911.00172,https://doi.org/10.48550/arxiv.2205.13792}. 

These ensemble weights are updated for every incoming token, although in separate experiments we observe that  we find that cluster assignments (and in effect, ensemble weights) can be fixed for the second half of a document with no drop in performance; this can further speedup inference.

We find that, in practice, the performance of our models is robust to even top-2 or top-4 experts, meaning that the inference costs of the language model are equivalent to a much smaller LM. We perform an empirical study of inference variations in \S\ref{sec:sparsity_analysis}.

\subsection{Comparing to Dense Training} 
\label{sec:comparing_dense}

Dense LMs are typically trained using hundreds or thousands of concurrent GPUs, all of which synchronize gradients each update.  For example, OPT-175B \citep{opt} was trained on 992 80GB A100 GPUs, and PaLM-540B \citep{palm} was trained on 6144 TPU v4 chips. \cbtm improves training efficiency by reducing communication overhead, as only GPUs training the same ELM must communicate. Furthermore, the chance of a GPU failure can grow considerably with the number of GPUs. \cbtm improves the resiliency of distributed training, since a single GPU failure only delays training for a single ELM, whereas in dense training, a single GPU failure afflicts training on all other GPUs.  \cbtm also makes training large LMs more feasible on shared GPU clusters, since it effectively decomposes training into smaller jobs which can run asynchronously. This makes job scheduling more efficient by reducing the number of GPUs that need to be allocated simultaneously.

\begin{figure*}[t!]
    \centering
    \includegraphics[width=\textwidth]{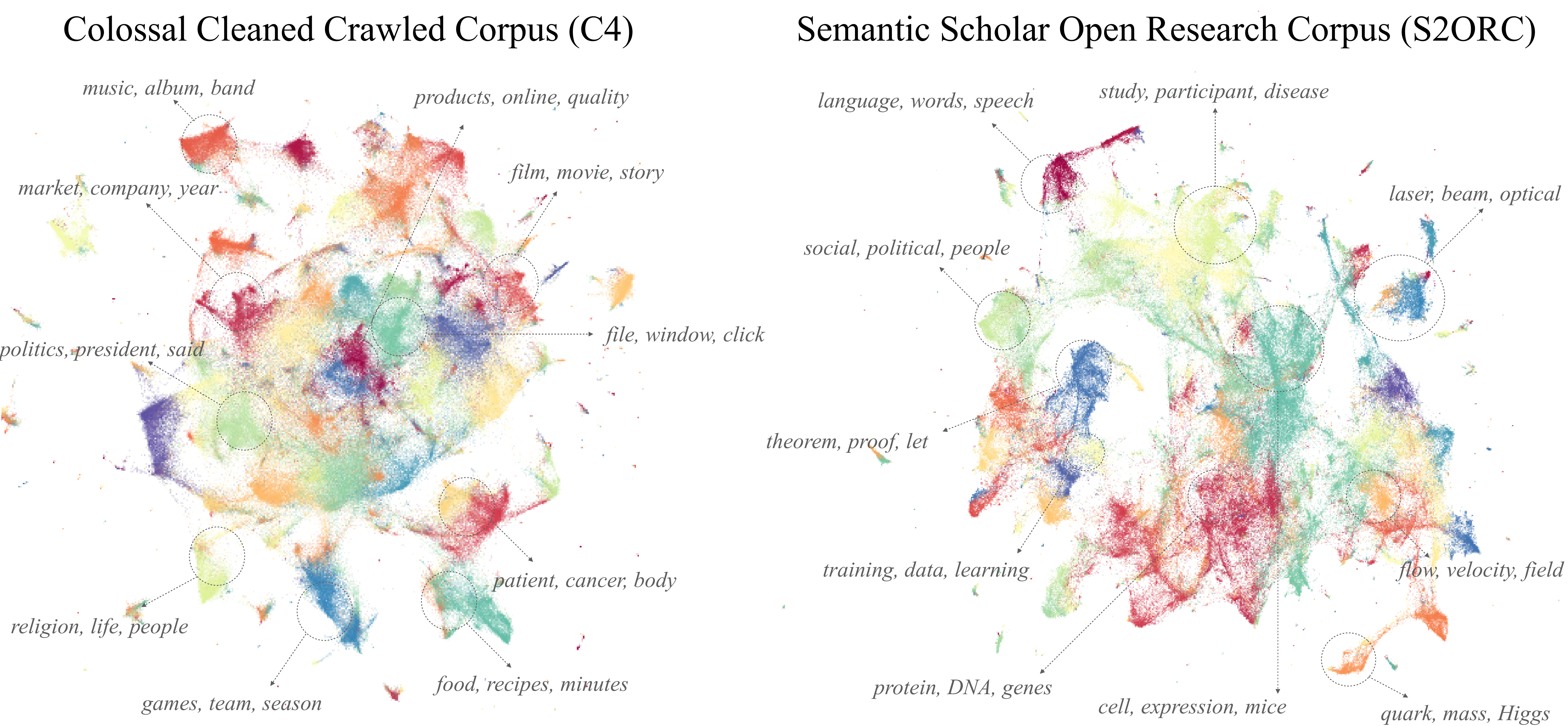} \vspace{-1em}
    \caption{\textbf{We train and evaluate on two large text corpora (\S\ref{sec:data}).} C4 (left; \citealt{t5}) 
 and S2ORC (right; \citealt{s2orc}) are diverse and contain many different clusters of text, indicated by these UMAP visualizations of 400K random documents in each corpus, colored with 32 automatically discovered and annotated clusters. See \S\ref{sec:analysis_specialze} for the description of our clustering and annotation procedure, and Figure~\ref{fig:c4} in the appendix for annotations of all clusters in these plots. }
    \label{fig:umaps}
\end{figure*}

\subsection{Comparing to BTM \citep{btm}}

Our method addresses several limitations of the training and inference techniques proposed by \citet{btm}.

First, BTM is limited to training data with metadata which can be used to determine its domains. Typical LM corpora, including C4 \citep{t5} and the Pile \citep{pile}, are sourced from the Internet without retaining document provenance at collection time, and are infeasible to label manually. Also, the optimal number of experts for a fixed corpus size, model architecture, and budget remains unknown, and is difficult to explore with metadata-based domains, since they cannot be easily merged or divided. \cbtm broadens the applicability of BTM to arbitrary datasets.

Moreover, BTM inference follows the \emph{cached prior} method introduced by \citet{gururangan-etal-2022-demix}, where the ensemble weights are estimated using Bayes' rule on additional held out data, and the prior $P(D=j)$ is estimated with an exponential moving average over sequences of posterior estimates that require forward passes on experts. This estimate is then fixed during test data evaluation. 

With \cbtm, we route based only on the current context. Thus, no additional data or forward passes through the experts are needed to estimate ensemble weights, nor do we need to assume that adjacent documents in the test set come from the same distribution. This also implies that the parameter averaging technique of \citet{btm} is not well suited to our setting, as it requires fixing the weights assigned to each expert for a set of evaluation documents. Future work may explore merging expert parameters for each context during inference.


\subsection{Comparing to Mixture-of-Experts (MoE)}
\label{sec:moe_comparison_summary}

Like MoE models \citep[e.g.,][]{https://doi.org/10.48550/arxiv.2209.01667}, \cbtm allows for efficient scaling of large LMs while keeping inference costs manageable. However, \cbtm routes sequences (instead of tokens) using offline balanced clustering (instead of online load balancing) with no shared parameters between experts. This eliminates effectively all complexities associated with balancing expert utilization \citep{baselayers}, avoids expensive all-to-all operations between experts \citep{artetxe2021efficient}, and naturally leads to interpretable expert specialization to domains of the training corpus.  In \S\ref{sec:moe_comparison}, we compare directly to MoE baselines trained with sparse upcycling  \citep{https://doi.org/10.48550/arxiv.2212.05055}, which initializes the MoE with a dense checkpoint, mirroring how \cbtm initializes ELMs.

\section{Experimental Setup}
\label{sec:experiments}

We design a set of experiments to study \cbtm on two large corpora (Figure \ref{fig:umaps}) selected to be distinct from  the corpus used to train our seed OPT model, and report perplexity on held out data from each corpus. 


\subsection{Data} \label{sec:data} 


\paragraph{C4 \citep{t5}} C4 is a publicly available distribution of a Common Crawl snapshot on Huggingface datasets.\footnote{\url{https://huggingface.co/datasets/c4}} We use the \emph{no blocklist} version (\texttt{en.noblocklist}) to train on a dataset that is out of distribution to our seed (OPT) pretraining corpus. C4 consists of 393M documents totaling 220B BPE tokens. We train on up to 168B tokens.

\paragraph{S2ORC \citep{s2orc}} The Semantic Scholar Research Open Corpus (S2ORC) is a publicly available corpus of full-text academic papers from the Semantic Scholar.\footnote{\url{https://allenai.org/data/s2orc}} The corpus spans 20 fields of study (e.g., Biology, Computer Science, Art), and contains 16M documents, totaling 87B BPE tokens. We train on up to 168B tokens over multiple epochs.\footnote{While it is not common to train large LMs for multiple epochs, we do not observe overfitting in any of our experiments, consistent with other studies that train LMs on academic literature for multiple epochs \citep{https://doi.org/10.48550/arxiv.2211.09085}.}

\paragraph{Evaluation data} For all experiments, we report language modeling perplexity on 200 randomly-sampled held out documents. Because S2ORC does not come with pre-defined validation data, we create a validation corpus by sampling an equal number of documents from each field of study.

\subsection{Experimental Setup}
\label{sec:experimental_setup}

\paragraph{Clustering the data} We segment each corpus using  balanced $k$-means clustering for $k \in $ \{2, 4, 8, 16, 32, 64, 128\} (\S\ref{sec:cbtm_iteration}). To train the clustering models, we first embed all data with a tf-idf vectorizer using scikit-learn,\footnote{\url{https://scikit-learn.org/}} with minimal assumptions: we only remove stop-words from a fixed lexicon and replace numbers with a dummy token. We then reduce the dimensionality of the resulting embeddings; we perform truncated SVD  with 100 dimensions, then normalize the vector by removing its mean and scaling to unit variance, which we observed in initial experiments improved the clustering quality. Finally, these representations are clustered using a custom Pytorch implementation.\footnote{\url{https://github.com/kernelmachine/balanced-kmeans}}  We present learned clusters and visualizations in Figure \ref{fig:umaps} and Figure~\ref{fig:c4} (in the appendix). We use a single shard of each training corpus (384K documents for C4, 155K documents for S2ORC) to train the clustering model and its embedder. No evaluation data is used in this process.

\paragraph{Seed LM} As LMs trained on diverse corpora make for better seeds \citep{btm}, we use pretrained OPT language models \citep{opt} as our seed for all experiments.

\paragraph{Model hyperparameters} We use the OPT architecture implemented in Metaseq \citep{opt}. We use OPT-1.3B for the initial set of experiments, and replicate our experiments with OPT-6.7B. Following \citealt{opt}, we use the GPT-2 vocabulary of 50,257 BPE types \citep{radfordlanguage}, and train with 2,048-token sequences, across document boundaries. We prepend a beginning-of-document token to each document. We set dropout to 0.1 for all parameters except those of the embedding layer. 

\paragraph{Training hyperparameters}  For all models, we fix the learning rate to that used during OPT pretraining (2e-4 for 1.3B parameter models; 1.2e-4 for 6.7B parameter models; \citealt{opt}) using a linear decay learning rate schedule to zero (with no warmup), which we found to work well for most settings after a grid search of fastest learning rates that avoided divergence.  We use a batch size of 8 for each GPU, and train with \texttt{fp16} and fully-sharded data-parallel \citep{artetxe2021efficient}. We train on NVIDIA V100 32GB GPUs. All models are trained with Metaseq \citep{opt}.  For a given number of clusters $k$ and total GPU budget $n$, each ELM is allocated $n/k$ GPUs, keeping the total effective number of FLOPs fixed across models exposed to the same number of tokens. See \S\ref{sec:comparing_flop_counts} for more details.

\paragraph{Scaling} We train for a total of 10K steps in each run; to expose the model to more tokens, we increase the total GPU budget proportionally, up to 64 GPUs. We simulate larger budgets, up to 1024 GPUs, by increasing gradient accumulation steps with 64 GPUs. This method of scaling increases the model's effective batch size for the same number of steps, and maintains near constant run-times across our many experiments. This experimental setup also means that as the number of clusters increases, the overall set of ELMs is exposed to more data with less simultaneous computation among GPUs. 


Other ways of training on more data (e.g., by keeping total batch size fixed and increasing step count) may yield different results. The best batch size and learning rate combinations for training language models are likely specific to a variety factors, including the model size, dataset, and total compute available \citep{https://doi.org/10.48550/arxiv.1811.03600, https://doi.org/10.48550/arxiv.1812.06162, NEURIPS2021_8df7c2e3}. In preliminary experiments, we found that expert models benefit from faster learning rates and larger batch sizes. Given a sufficiently large batch size, experts are robust to a variety of learning rates. Our larger budget experiments might benefit from higher learning rates, but we leave further tuning for future work.

\paragraph{Inference} One of the key hyperparameters for inference is the temperature $T$ (Equation 3),  which governs the sharpness of the probability distribution over experts for a given context. We find that setting $T$=0.1 works well for most settings (see \S\ref{sec:appendix_sparsity} for more details). We also compute the nearest cluster centers for every incoming context, regardless of how stable the cluster assignments already are for a document. However, we find that these assignments can be fixed for the second half of a document with no drop in perplexity; this can further speedup inference. The other important hyperparameter is the top-k value, which sparsifies the probability distribution over experts. For our core experiments in \S\ref{sec:core_results}, we set top-k to the total number of experts we have trained for each model. We explore the effect of enforcing sparsity with lower top-k values in \S\ref{sec:sparsity_analysis}. 

\paragraph{Baselines} 
In our primary experiments (\S\ref{sec:results}), we compare with a strong dense baseline (i.e., our 1-cluster model) following OPT pretraining. We also progressively increase the number of clusters we train with for a fixed number of tokens. In subsequent experiments (\S\ref{sec:moe_comparison}), we compare to MoE language models initialized from a dense checkpoint.

\subsection{Making Fair Model Comparisons} 
\label{sec:fair_comparisons}

We follow the recommendations of \citet{https://doi.org/10.48550/arxiv.2110.12894} and report results with multiple cost metrics, and detail our choices here.
When comparing model training budgets, we are primarily concerned with the true monetary cost of model training, which is typically billed in direct proportion to GPU-time. Model inference comparisons have two main considerations: monetary cost incurred by the model deployer, again measured in GPU-time, and latency for end-users, or wall-clock time (i.e., how slow a model inference is for an end-user).

We explicitly do \emph{not} compare or match the number of model parameters during training, which has minimal bearing on the cost of model training separately from its influence on GPU-time. The number of training parameters is a particularly misleading cost measure that is unsuitable for sparse models, since they can maintain the FLOPs and inference speed of dense models despite training many more parameters \citep{https://doi.org/10.48550/arxiv.2110.12894}.

\begin{figure}[t!]
    \centering
    \includegraphics[width=\columnwidth]{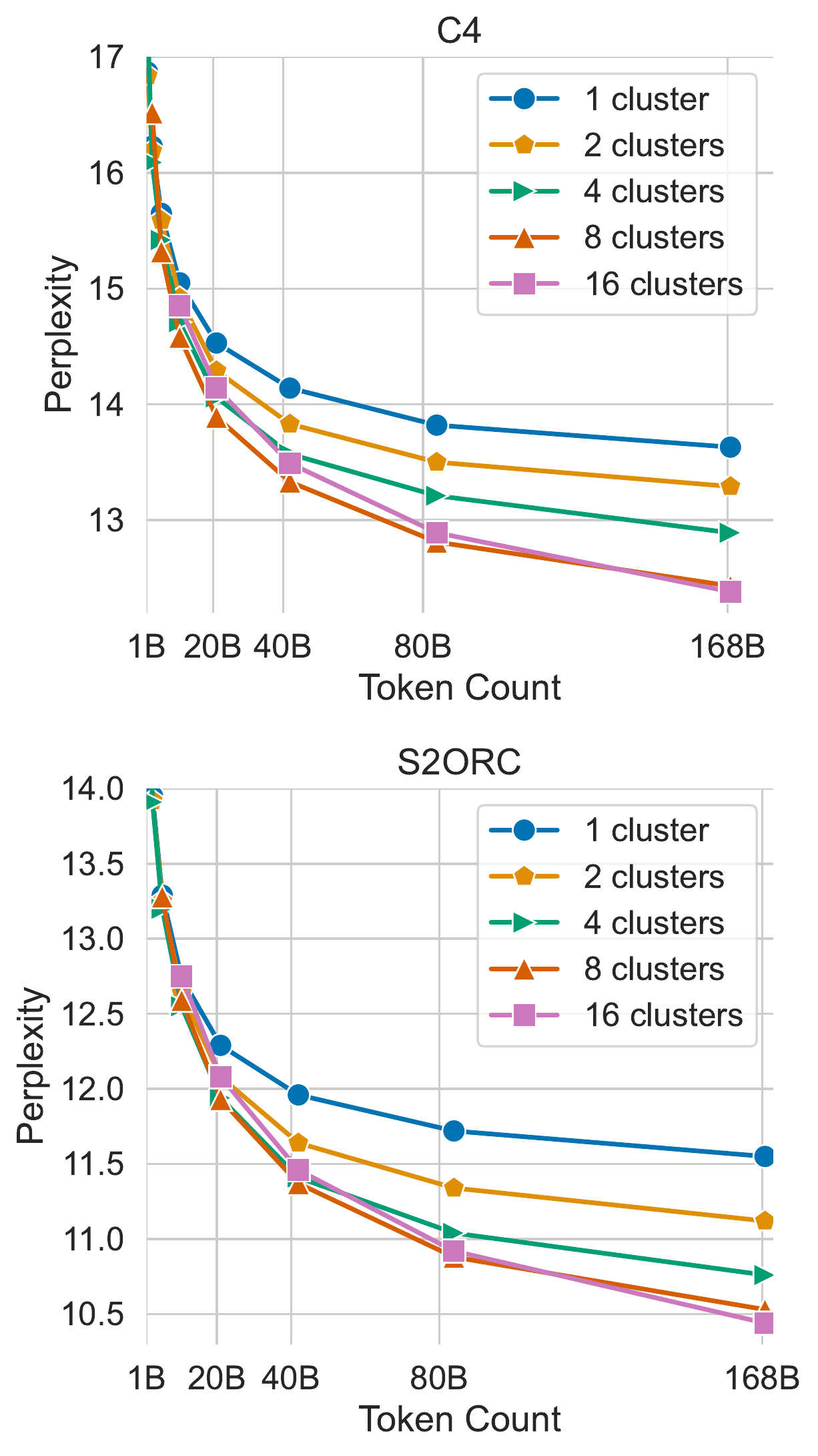} 
    \vspace{-2em}
    \caption{\textbf{Increasing cluster count in \cbtm improves language modeling performance for a fixed compute budget (\S\ref{sec:core_results}).} Performance of \elm{}s trained with \cbtm as a function of the total number of tokens trained on, which, in our experiments, equalizes FLOP count. Training with more than one cluster always outperforms the compute-matched, single cluster dense model, and we observe improving performance (and in \S\ref{sec:time_comparison}, faster updates) as we increase the number of clusters.} 
    \label{fig:baseresults}
\end{figure}

\paragraph{Training GPU-time} Assuming fixed hardware, GPU-time during model training is determined mostly by FLOPs and inter-machine communications. However, prior work typically only FLOP-matches, ignoring the additional inter-GPU communications incurred by some models (e.g., MoE) that increase training costs. Ideally, our comparisons could directly fix GPU-time. This is challenging in practice, as even identical computations on the same GPU node at different times can vary wildly in speed due factors like  temperature, other activity on the node, or the quality of GPU interconnect. To maintain consistency and fairness despite these confounds, our results compare FLOP-matched models with the same training data budget over the same number of updates (\S\ref{sec:core_results}), but also report the speed of training for each FLOP-matched model (\S\ref{sec:time_comparison}). This allows us to disentangle and accurately reflect multiple cost metrics of training. Since, in our experiments, models exposed to the same number of tokens incur the same number of FLOPs, we use training data size as a more interpretable measurement of the overall training budget (see \S\ref{sec:comparing_flop_counts} for more details). 

\paragraph{Inference GPU-time} Inference GPU-time is also primarily the result of FLOPs and communication costs. Since communication during inference is minimal,  we compare FLOPs via inference parameters (\S\ref{sec:sparsity_analysis}). We do not account for the FLOPs of the \cbtm router, which varies based on the clustering approach, and is relatively negligible.

\paragraph{Inference latency} FLOPs is not an ideal metric for inference latency of our models, because \cbtm allows for parallel inference across ELMs. This means that if ELMs share the same architecture (e.g., OPT-1.3B), inference latency is  always equivalent to that of a single ELM, regardless of the number of experts active.  However, inference latency may be quite different between model architectures (e.g., OPT-1.3B and OPT-6.7B); we discuss this further in \S\ref{sec:zflops}. As with inference GPU-time, we do not consider the latency of the \cbtm router.








\begin{figure}[t!]
    \centering
    \includegraphics[width=\columnwidth]{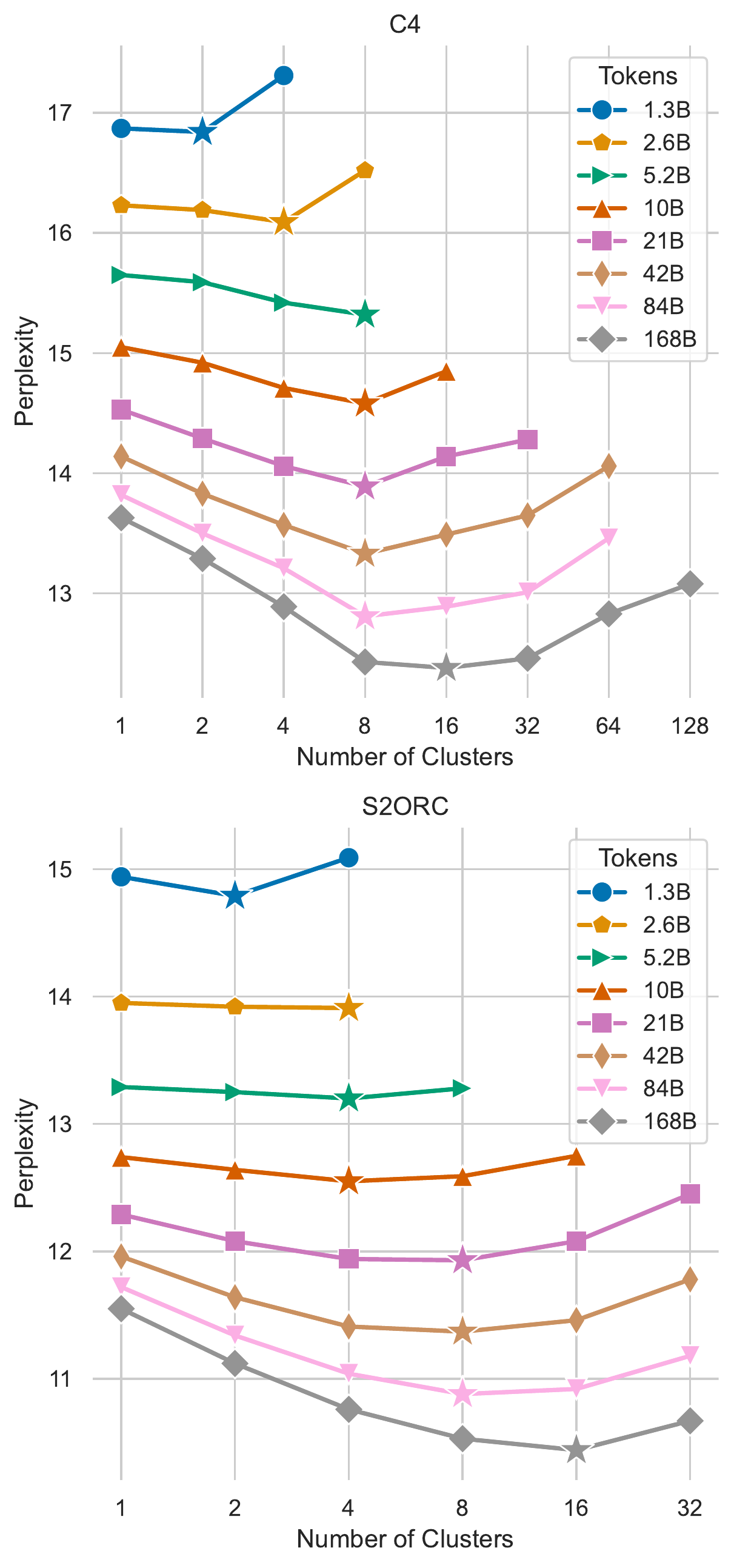} 
    \vspace{-2em}
    \caption{\textbf{There exists an optimal cluster count for each compute budget (\S\ref{sec:core_results}).} The optimal cluster count increases as one increases the compute budget, but using too many clusters \emph{without} sufficiently increasing compute can degrade performance. For both C4 and S2ORC, 16 clusters gets the best performance at the highest budget (168B tokens), although higher cluster counts still outperform the 1-cluster (dense) model ($x$ = 1 in this graph).}
    \label{fig:optimalclusters}
\end{figure}

\section{Language Modeling Results}
\label{sec:results}


We begin with a set of experiments in which we train LMs with \cbtm on datasets from \S\ref{sec:data}. We are interested in measuring how performance changes as we increase overall compute.
We first compare models against training costs: \emph{total training tokens} (\S\ref{sec:core_results}) and \emph{training time} (\S\ref{sec:time_comparison}). Then, in \S\ref{sec:sparsity_analysis}, we compare model performance along an axis of inference costs: the \emph{total parameter count at inference time}. Finally, in \S\ref{sec:zflops} we compare model performance by fixing both training and inference costs. Across all computational budgets, \cbtm provides substantial benefits over dense training, and performance improvements increase as the total compute grows.

\subsection{Controlling for Total Training Tokens}
\label{sec:core_results}

First, we compare model performance controlling for overall training data size (or equivalently, training FLOPs; \S\ref{sec:fair_comparisons}). Figure \ref{fig:baseresults} shows evaluation perplexity on C4 and S2ORC with up to 16 clusters. Training on more than one cluster always outperforms training with a single cluster (i.e., a dense model). As the amount of training data grows, the gap between our models and the dense one widens, indicating that experts make better use of larger training datasets, possibly due to their increased specialization. These results suggest that as we increase the amount of data available, \cbtm benefits from more clusters.

However, Figure \ref{fig:optimalclusters} shows that there exists an optimal cluster count for each token budget that we consider. Each number of clusters has a budget range in which they are optimal, and the optimum smoothly progresses from smaller to larger cluster counts as we increase the training data size. If we increase the cluster count past the optimum, each expert has an insufficient share of the data, resulting in worse performance. 

Nevertheless, we observe that using more clusters than optimal for the highest token budget settings still outperforms the dense model. Since it is cheaper to train with more clusters for a fixed training data size due to parallelism, it may be preferable in some settings to train with a large number of clusters despite their less-than-optimal performance. Based on the trends we observe at this scale, we expect that higher cluster counts would become optimal as we scale the training data size even further. 

The consistency of our results on C4 and S2ORC suggests that these general trends may be widely applicable to many datasets.  However, the optimal number of clusters for a given computational budget is likely dataset specific. Future work may explore relationships between dataset features and the optimal cluster count.

\begin{figure}[t!]
    \centering
    \includegraphics[width=\columnwidth]{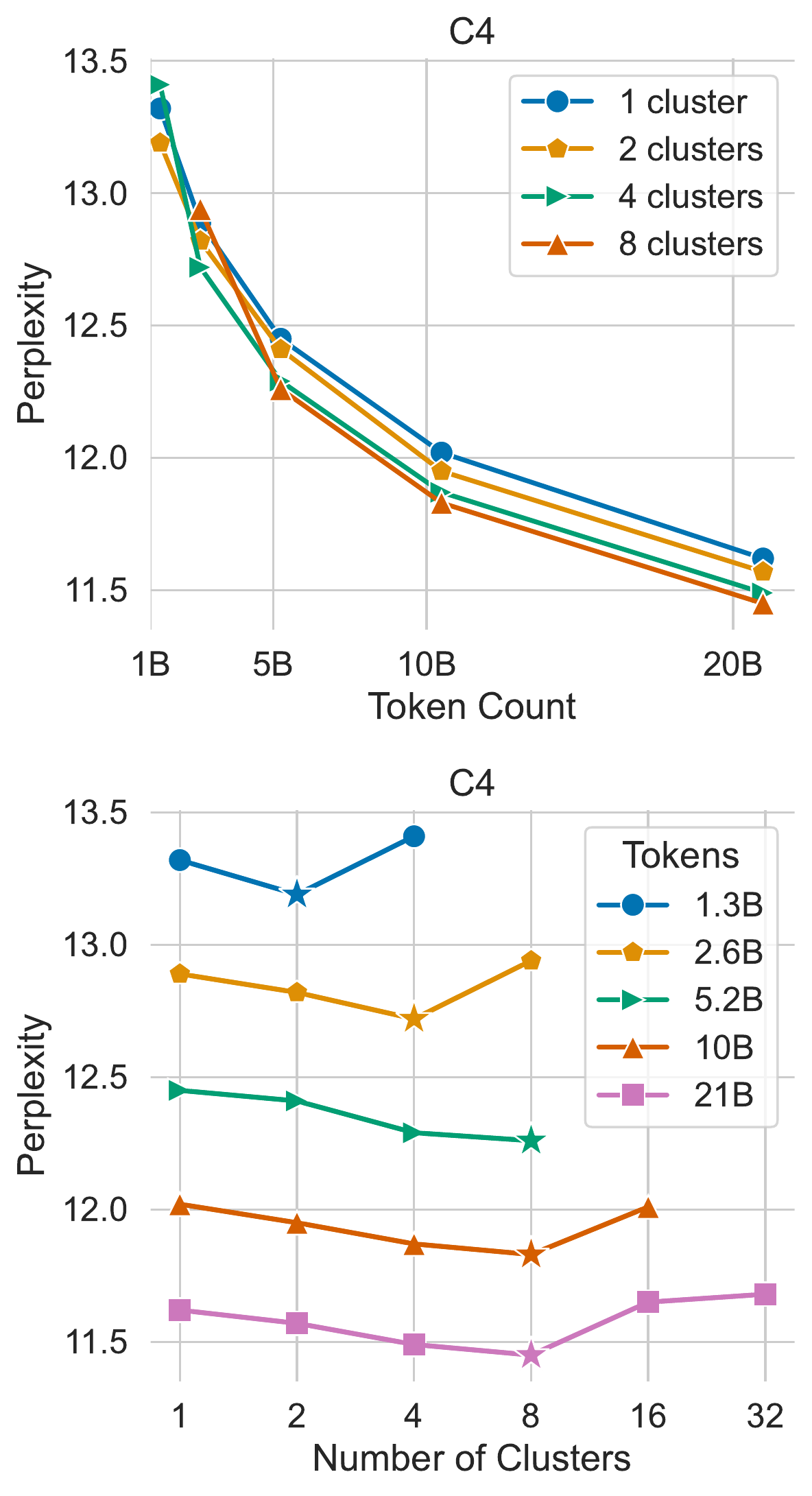} 
    \vspace{-2em}
    \caption{\textbf{Our results are consistent even as we increase expert size to 6.7B parameters (\S\ref{sec:core_results}).} 8 clusters is optimal at 21B tokens, as it is for the 1.3B parameter ELMs. However, the gaps between these models are smaller, due to the substantial increase in pretraining FLOPs for the OPT-6.7B checkpoint. }
    \label{fig:67b_results}
\end{figure}

These trends are consistent as we increase the size of our experts to 6.7B parameters (Figure \ref{fig:67b_results}), although the gaps between our baselines reduce, likely due to the substantial increase in pretraining FLOPs for OPT-6.7B.\footnote{OPT-6.7B was pretrained for 1.83 ZFLOPs, while the OPT-1.3B was trained for 0.34 ZFLOPS.}

\subsection{Comparing Training Time}
\label{sec:time_comparison}



\begin{figure}[t!]
    \centering
    \includegraphics[width=\columnwidth]{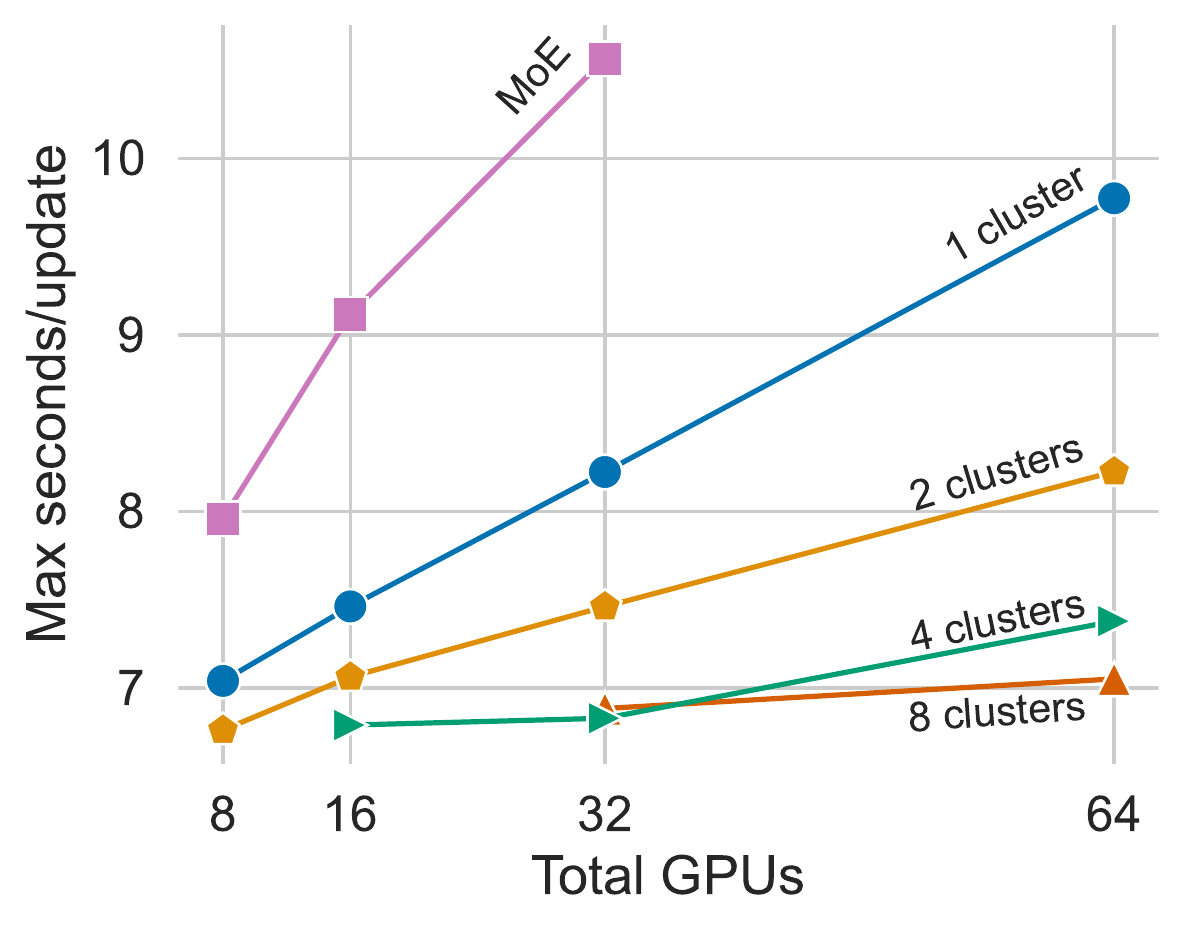} 
    \vspace{-1em}
    \caption{\textbf{Models trained with more clusters have faster updates as we increase the total compute (\S\ref{sec:time_comparison}).} We display the maximum seconds-per-update for \cbtm and MoE models with varying GPU counts (across all experts). Under fixed compute, training with more clusters uses fewer GPUs per expert, and \cbtm avoids communication between experts, resulting in faster updates. On the other hand, MoE models are much slower to train, due to extensive communication between experts (\S\ref{sec:moe_core_results}), as well as additional FLOPs from top-2 routing \citep{artetxe2021efficient}.}
    
    \label{fig:ups}
\end{figure}

\begin{figure}[t!]
    \centering
    \includegraphics[width=\columnwidth]{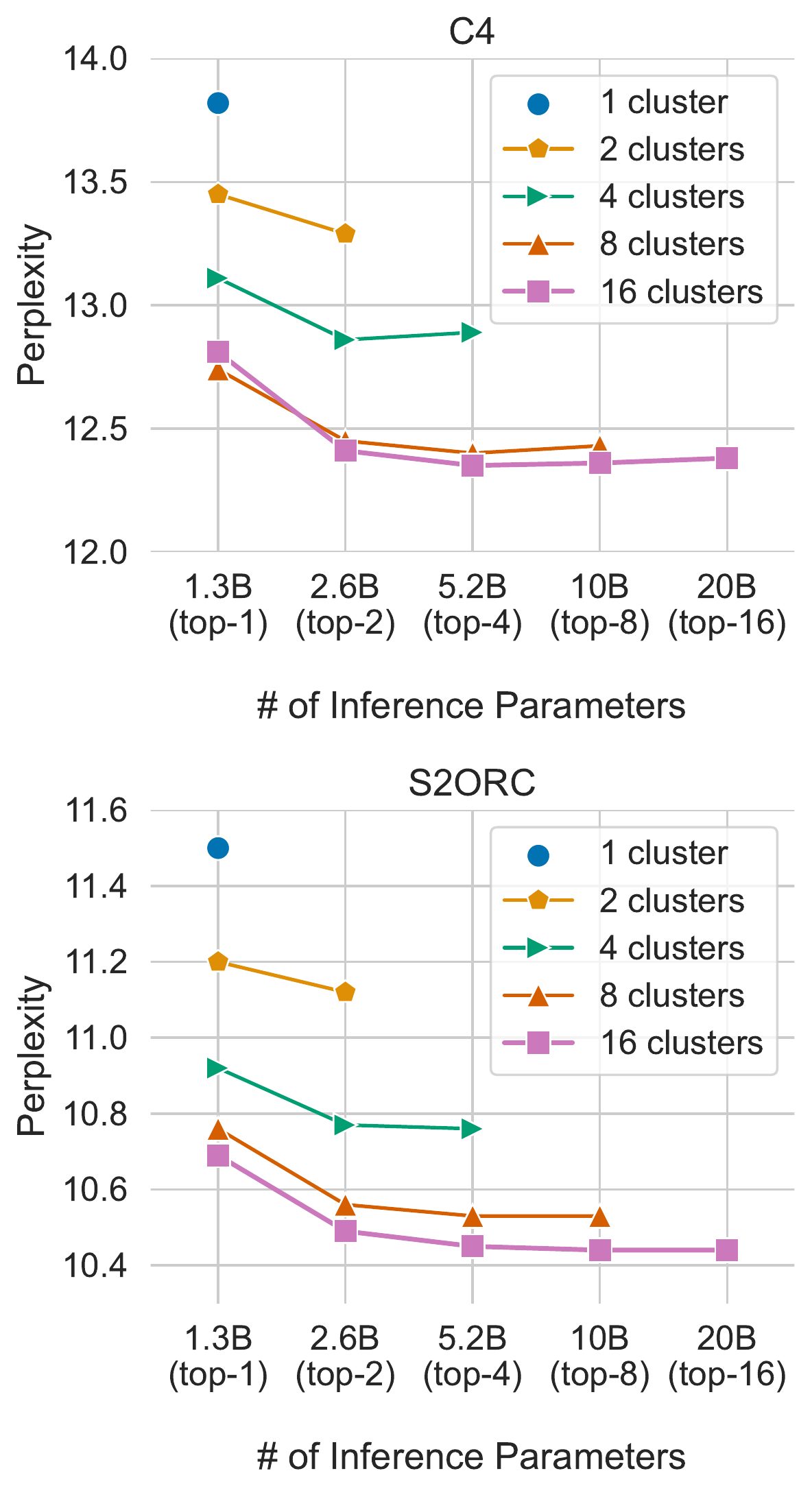} 
    \vspace{-1em}
    \caption{\textbf{Sparse top-$k$ inference performance at 168B token budget (\S\ref{sec:sparsity_analysis}).} ELMs perform well even with heavily sparsified inference. Top-1 inference substantially outperforms the densely trained baseline at no additional inference cost, and top-2 and top-4 inference performs comparably to (and is sometimes slightly better than) activating all experts. In our setup, inference parameters are proportional to inference FLOP count (\S\ref{sec:fair_comparisons}).}
    \label{fig:sparsity}
\end{figure}

 Now, we turn to comparing our models based on training times. We measure the speed of training each model with the maximum seconds-per-update for each training run.\footnote{Other measures of seconds-per-update (e.g., average, median) tend to be noisy, due to factors such as dataloading and bursty GPU activity.} For \cbtm models with more than one cluster, we use the maximum seconds-per-update across all experts. To make our comparisons fair, we only compare the training times of models that have the same effective batch size (\S\ref{sec:fair_comparisons}). Our results are displayed in Figure \ref{fig:ups}. As we increase the number of clusters and training data size, the update speed for \cbtm \emph{increases}, since models with higher cluster counts use fewer GPUs per expert under a fixed budget, and there is no communication between experts. This suggests that \cbtm models with more clusters can be exposed to more data for the same amount time as dense models.

As discussed in \S\ref{sec:comparing_dense}, \cbtm also provides important practical speedups when training large LMs at scale. \cbtm divides large compute budgets among many models, such that we can train on 168B tokens with only 8 GPUs per expert in the 128-cluster setting. On shared multi-node clusters, allocating many smaller jobs incurs shorter cumulative wait times than a single, large synchronous job, since they can make more efficient use of shared resources, and run on short-lived, idle nodes \citep{lofi}. Furthermore, large LM training is prone to node failures, gradient spikes, and other unexpected behaviors \citep{opt}. With dense models, when one node fails, all nodes must restart due to synchronization. With \cbtm, experts are trained independently; if a node fails, only the corresponding expert needs to be restarted, and all other experts are unaffected.

\subsection{Controlling for Inference Costs via Parameter Count}
\label{sec:sparsity_analysis}

\begin{figure}[t!]
    \centering
    \includegraphics[width=\columnwidth]{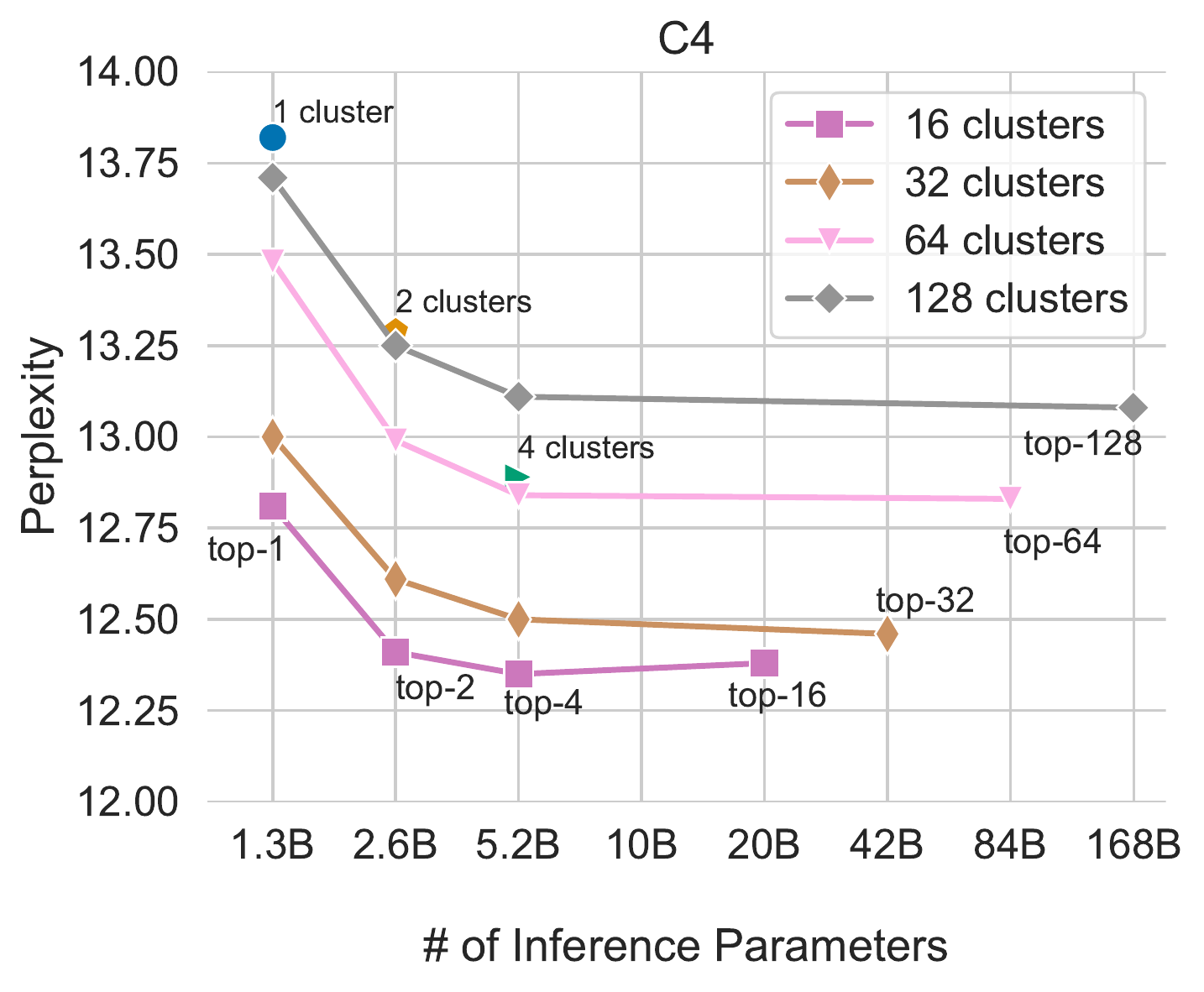} 
    \vspace{-2em}
    \caption{\textbf{Train large, then sparsify (\S\ref{sec:sparsity_analysis}).} Despite using more than the optimal cluster count for the 168B token budget, sparsifying the 32, 64, and 128-cluster models with top-1, top-2, or top-4 experts is usually better than training 1-, 2-, or 4-cluster models.}
    \label{fig:sparsity_larger}
\end{figure}


Comparing models with just training budgets ignores the fact that \cbtm inference GPU-time costs grow as we increase the number of clusters, since we train more parameters. To mitigate these costs, we can use the top-k function (Equation \ref{eq:topk}) to dynamically use a subset of experts for each incoming context during evaluation (\S\ref{sec:cbtm_inference}). Next, we study the effect of inference parameter count on model performance.  We focus on the largest training budget (i.e., 168B tokens) for these experiments.

 Results (Figure~\ref{fig:sparsity}) show that despite training many more parameters, training \cbtm with many clusters and then using only the top-1 expert still outperforms the dense model. Further, using the top-2 or top-4 experts yields comparable performance to activating all experts. Sometimes we observe that sparsifying can even slightly \emph{improve} performance over using all experts (for example, see the 16 cluster model for C4 in Figure \ref{fig:sparsity}). We speculate that having all experts active may introduce interference effects from experts that are specialized to clusters unrelated to test-time contexts. 

Our results in Figure \ref{fig:sparsity_larger} suggest that sparsifying even larger expert models (i.e., those with more clusters than the optimal for a given token budget) is still highly effective. At the most extreme setting, using the top-1 expert for the 128 cluster model (using 0.7\% of total parameters at inference time for each context) still outperforms the dense model, and the top-4 expert model (3.1\% of total parameters) performs comparably to using all experts.

These results suggest that \cbtm results in a highly sparse LM, and that inference costs can be kept constant even as the number of experts grows, though additional experts can be added to further boost performance. 


\subsection{Comparing to a Larger Dense Model} 
\label{sec:zflops}



In our final comparison of this section, we consider both training and inference costs together. We compare a 6.7B 1-cluster (dense) model and \cbtm model with 1.3B parameter experts, which uses 16 clusters (optimal in our experiments from \S\ref{sec:core_results}) and top-4 inference, resulting in 5.2B inference parameters.  This \cbtm model has lower inference cost than the larger 6.7B parameter dense model (\S\ref{sec:fair_comparisons}). The former uses fewer inference parameters, incurring a smaller inference GPU-time cost, and has lower latency, comparable to that of a single 1.3B-parameter ELM.

We compare the FLOPs used to train each model. Following \citet{artetxe2021efficient}, we build continuous efficiency curves by interpolating between our empirical observations. Specifically, we calculate the speedup between our cluster expert models and dense model by interpolating between the discrete observations of perplexity values for a given empirical number of FLOPs.\footnote{See \S\ref{sec:perf_interpolation} for details on this interpolation.} Our goal is to identify the FLOP count necessary to achieve a particular perplexity value. If ELMs trained with \cbtm achieve the same perplexity as the dense model with half the FLOPs, we conclude that \cbtm achieves a 2$\times$ speedup.

\begin{figure}[t!]
    \centering
    \includegraphics[width=\columnwidth]{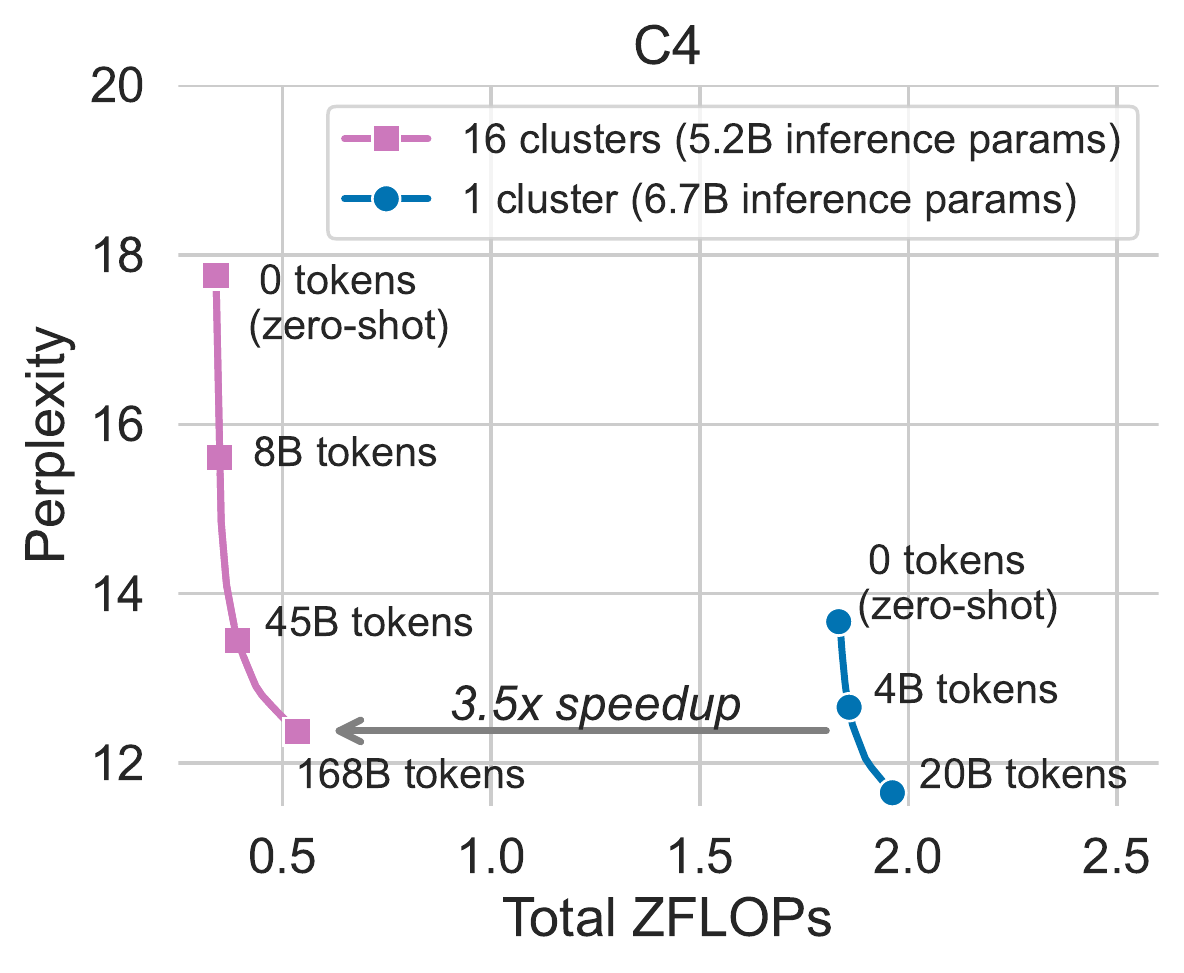}
    \caption{\textbf{Training with \cbtm is substantially more efficient than training a larger dense LM (\S\ref{sec:zflops}). }   We train a 16-cluster \cbtm model and use top-4 inference, resulting in a 5.2B parameter LM, and compare its performance with a 6.7B parameter dense LM. The \cbtm model at 168B tokens achieves the same perplexity on C4 as a 6.7B dense model with 3.5x fewer ZFLOPs. The total ZFLOPs includes the cost of pretraining the seed OPT checkpoints.}
    \label{fig:zflopcomparison}
\end{figure}

Our results are presented in Figure \ref{fig:zflopcomparison}. A smaller \cbtm model, exposed to 168B tokens of text, can achieve the same perplexity as the larger 6.7B dense model with 3.5$\times$ speedup. These speedup estimates are dependent on the amount of pretraining performed on each model.  Future work may perform these experiments  with larger models and many more ELMs.

\subsection{Summary} 
\label{sec:results_summary}

Our results demonstrate that controlling for a variety of different types of computational budget, \cbtm outperforms dense training in language modeling. Furthermore, we demonstrate that \cbtm results in an effective \emph{sparse} language model, where the top-1, top-2 and top-4 experts from models with at least 8 clusters significantly outperform 1-cluster, 2-cluster and 4-cluster models. These results suggest the possibility of  outperforming dense models by increasing margins, while keeping both training and inference costs fixed, as compute and the number of experts grow.


\begin{table*}[t!]
\centering
\small
\begin{tabular}{llrrrrrrrr}
& & \multicolumn{6}{c}{\it Few-shot Text Classification Accuracy (\%)} \\
\toprule
  & & \bf AGNews & \bf DBPedia & \bf SST-2 & \bf Amazon & \bf Phrasebank  & \bf Twitter  \\ 
&  $\downarrow$ Model (inference parameters) &  \it Topic & \it Topic & \it Sentiment & \it Sentiment & \it Sentiment & \it Hatespeech & \bf Average  \\

\midrule
& Random chance & 25.0 & 7.10 & 50.0 & 50.0 & 33.3 & 50.0 & 35.9  \\
& OPT (1.3B) &  42.9 & 57.2 & 72.8 & 81.3 & 72.5 & \bf 65.1  & 65.3 \\
& OPT (6.7B) & 51.9 & 58.9 & 77.0 & 83.8 & 76.4 & 39.6 & 64.6  \\
\midrule
\parbox[t]{3pt}{\multirow{5}{*}{\rotatebox[origin=c]{90}{\bf {\textsc{C4}}}}}  & 1-cluster (1.3B) & 47.4 & 61.1 & 80.2 & 80.7 & 66.6 & 60.9 & 66.2 \\
& 1-cluster (6.7B) & \bf 68.1 & 62.4 & 80.7 & \bf 84.9 & \bf 80.6 & 37.4  & 69.0 \\
&   16-cluster; top-1 (1.3B) &  47.1	& \bf 62.9	& 74.3 & 	79.1 &	72.9 &	56.4  & 65.4 \\
& 16-cluster; top-4 (5.2B)  & 49.3	& 62.3 &  80.0 &	81.3 &	  78.7 & 	61.3 & 68.8 \\
& 16-cluster; top-16 (20.8B) & 50.6 &	62.0 &	\bf 84.0 & 83.2 &	78.6 &	 61.7 & \bf 69.9 \\
\bottomrule
\end{tabular}
\caption{\textbf{\cbtm models outperform dense counterparts on downstream  text classification tasks (\S\ref{sec:downstream_task_results}).} We display accuracy of models from \S\ref{sec:results} on six text classification tasks, using eight demonstrations for each example and no additional fine-tuning. We report accuracy averaged over five random seeds. The 1- and 16-cluster models are trained on 168B tokens of C4 (i.e., our highest budget). The 16-cluster model, with top-4 or top-16 inference, always outperforms the 1-cluster model, and top-1 inference usually outperforms the 1-cluster model at no additional inference cost. We include average performance of models across tasks for readability.}
\label{tab:downstream_tasks}
\end{table*}

\section{Downstream Task Results}
\label{sec:downstream_tasks}

Do the trends from \S\ref{sec:results} extend to downstream settings? To begin to answer this question, we perform few-shot evaluation on six downstream text classification tasks. We indeed find that models trained with \cbtm outperform their dense counterparts in these settings.

\subsection{Experimental Setup}

\paragraph{Tasks} We experiment with six text classification tasks, spanning topic, sentiment, and hatespeech classification. Details of the datasets are in Appendix \ref{sec:appendix_downstream_tasks}.

\paragraph{Few-shot inference} We perform 8-shot evaluations. For each task, we randomly sample 8 examples with their labels from the train set, and prepend them as demonstrations for each test example. For \cbtm models, we estimate ensemble weights for each example by passing both the example and the demonstrations through our pretrained clusterer (\S\ref{sec:cbtm_inference}).  We calculate the probability of each label for the task under the model, and report accuracy by counting the proportion of test examples where the gold label has the highest probability. We report average accuracy over 5 random seeds. We leave careful analysis of \cbtm with varying numbers of demonstrations and few-shot inference techniques to future work.

\paragraph{Baselines} We compare the performance of 1- and 16-cluster \cbtm models trained on 168B tokens of C4 (i.e., our highest budget from \S\ref{sec:results}). For the 16-cluster model, we also perform top-1 and top-4 inference (\S\ref{sec:sparsity_analysis}). We additionally compare against a random baseline, the original OPT-1.3B and 6.7B models (without any additional training), and the 6.7B parameter 1-cluster model trained on 20B tokens of C4.

\subsection{Results}
\label{sec:downstream_task_results}


Our results in Table \ref{tab:downstream_tasks} show that the 16-cluster \cbtm model always outperforms the 1-cluster, 1.3B parameter baseline, sometimes dramatically. This aligns with our language modeling results (\S\ref{sec:core_results}). The 1-cluster model achieves lower accuracy than OPT-1.3B on some tasks despite additional training, suggesting that our models may suffer from catastrophic forgetting, since the C4 corpus is out-of-domain to OPT.

Nevertheless, the 16-cluster model outperforms OPT-1.3B on all tasks other than Twitter. Also, top-1 and top-4 inference matches or exceeds using all experts in some settings, consistent with our language modeling results in \S\ref{sec:sparsity_analysis}. We examine the clusters associated with the most likely experts for each task, and find that their top-terms are relevant to the task's domain (Table \ref{tab:topk_examples} in the appendix). This supports our hypothesis that \cbtm is able to leverage any part of the corpus which is in-domain to the test task, even if the training corpus as a whole might be sufficiently out-of-domain as to have a negative effect on performance. 

We then mirror the analysis in \S\ref{sec:zflops}, by comparing our 16-cluster models to 6.7B parameter dense models. First, we observe that our 1-cluster 6.7B model outperforms OPT-6.7B on all tasks except Twitter, possibly because this model has had less exposure to C4, and suffers from less catastrophic forgetting. Our 16-cluster model performs comparably to both 6.7B models, and on multiple tasks, our 16-cluster model \emph{outperforms} both 6.7B models, which have been trained with at least 3.5$\times$ more compute (\S\ref{sec:zflops}). With top-4 inference, \cbtm models activate even fewer parameters than the 6.7B parameter models, yet perform comparably. These results corroborate our findings in \S\ref{sec:zflops} that compared to larger dense models, models trained with \cbtm have more training efficiency and lower inference latency, and result in comparable or better performance.

In separate experiments, we observe that routing examples to experts based on their performance on few shot examples, rather their clusters, results in even better downstream task performance with \cbtm models. This is likely because few-shot performance depends on factors such as example order, label distributions, and the quality of the demonstrations, which not necessarily tied to the domain of the task \citep{min-etal-2022-rethinking, lu-etal-2022-fantastically}. We analyze this finding further in \S\ref{sec:appendix_downstream_tasks}, and leave more careful development of routing protocols for downstream tasks to future work.

\subsection{Summary}

We demonstrate that, consistent with the language modeling results in \S\ref{sec:core_results}, \cbtm improves downstream performance on a variety of few-shot text classification tasks. \cbtm models consistently outperform dense 1-cluster baselines, and usually outperform the original OPT models, despite being trained on an out-of-domain corpus. We also find that top-$k$ activation reduces inference costs with negligible effects on downstream task performance. \cbtm models perform comparably to larger, 6.7B OPT and 1-cluster dense baseline models, despite being trained with 3.5x less compute, and even when activating fewer inference parameters.

\section{Comparing to Mixture-of-Experts}
\label{sec:moe_comparison}

Finally, we compare \cbtm against an alternative sparse LM, a mixture-of-experts (MoE) which learns a routing between tokens and feedforward experts in the transformer \citep{lepikhin2020gshard, fedus2021switch}. As discussed in \S\ref{sec:moe_comparison_summary}, \cbtm is substantially simpler than MoE.

\subsection{Sparse Upcycling}

To mirror \cbtm seed initialization, we initialize our MoE with a dense checkpoint. We use the \emph{sparse upcycling} technique from \citet{https://doi.org/10.48550/arxiv.2212.05055}. Upcycling a dense model into an MoE with $k$ experts entails initializing shared parameters (e.g., attention and embedding layers) and $k$ expert parameters (e.g., every other feedforward layer) from a dense checkpoint, and initializing new parameters for the router. Then the model is simply trained as an MoE. Here, we use top-2, token-level routing \citep{lepikhin2020gshard}.

\subsection{Experimental Setup} 

\paragraph{Hyperparameters} We train an MoE with sparse upcycling on C4, starting from OPT-1.3B and using the same general experimental setup detailed in \S\ref{sec:experimental_setup}. We follow the settings from \citet{https://doi.org/10.48550/arxiv.2212.05055} as closely as possible. We conducted experiments with 8, 16, 32, 64, and 128 experts for each compute budget. 8 and 16 experts are similar to, but slightly worse than, 32 experts; 64 experts and 128 experts consistently have exploding losses, and the few which successfully train are also similar to but slightly worse than 32 experts. In general, we find that both large expert count (and higher compute budgets) result in sparse upcycling training instability. 

We use 32 experts in our MoE, a capacity factor of 2, and continue training without resetting the optimizer from that used during OPT pretraining. We set all hyperparameters to be the same as our \cbtm models (\S\ref{sec:experimental_setup}), except that we use a peak learning rate of 2e-5, which we found to be the highest learning rate that that did not result in divergence after a sweep. We release our code for sparse upcycling, implemented in Fairseq \citep{https://doi.org/10.48550/arxiv.1904.01038}, publicly.\footnote{\url{https://github.com/kernelmachine/moe-fairseq}}

\paragraph{Baselines} We compare the 32-expert MoE LM to 1-cluster (i.e., dense) and 16-cluster \cbtm models.

\begin{figure}[!t]
    \centering
    \includegraphics[width=\columnwidth]{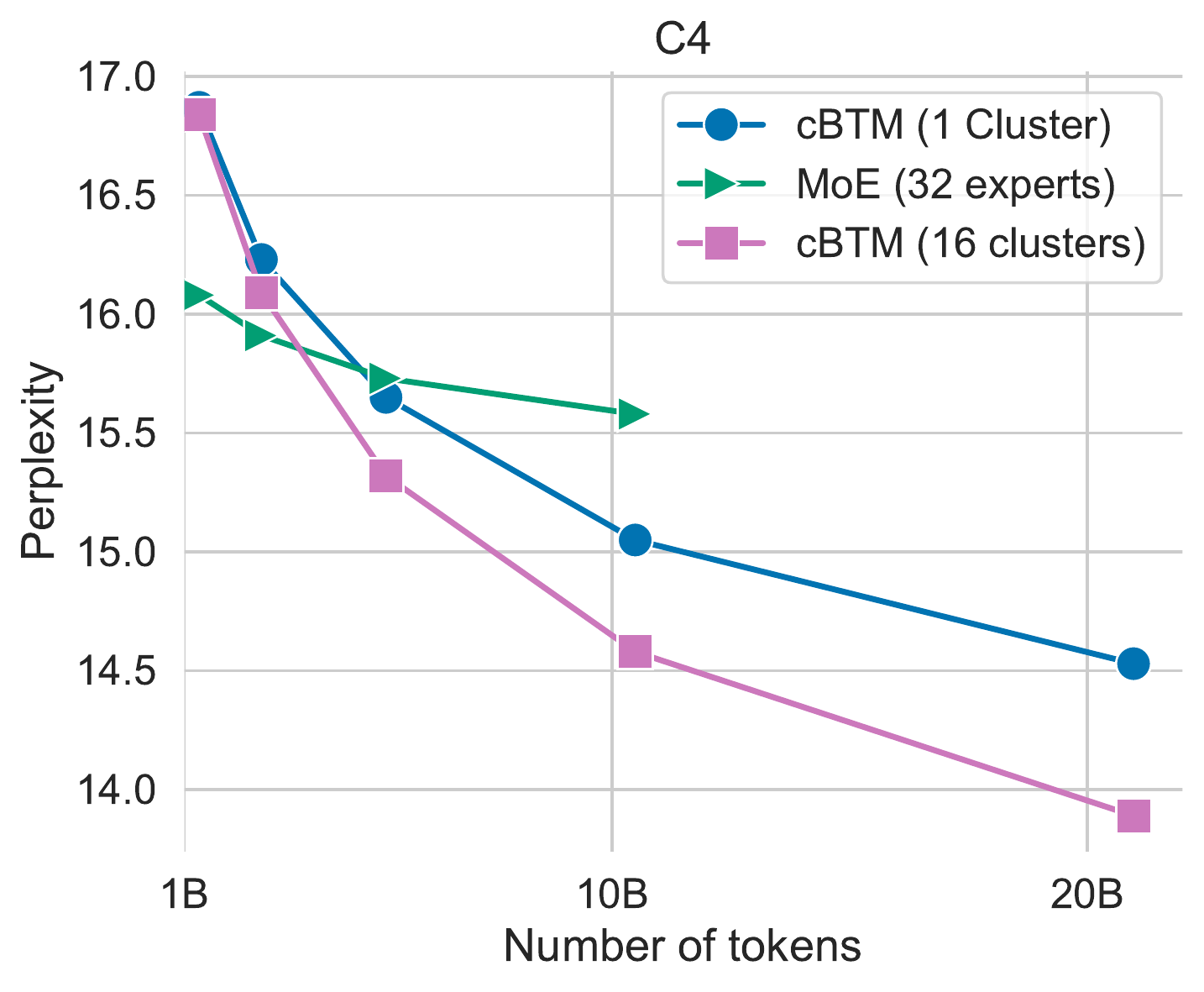} \vspace{-2em}
    \caption{\textbf{MoE underperforms \cbtm  (\S\ref{sec:moe_core_results}).} We compare a 32-expert MoE with top-2 routing \citep{lepikhin2020gshard} trained with sparse-upcycling \citep{https://doi.org/10.48550/arxiv.2212.05055}. While the MoE outperforms \cbtm models with 16 experts at small budgets, it fails at larger budgets, even under-performing the dense model. We speculate this could be due to distribution shifts after pretraining, which might increase the instability of upcycling.}
    \label{fig:sparse_upcycling}
\end{figure}

\subsection{Results}
\label{sec:moe_core_results}

The MoE expends more FLOPs than the other models due to the additional feedforward layer at every other transformer block for top-2 routing, as well as the routing projection \citep{artetxe2021efficient}. For clarity and consistency, we update-match and separately report GPU-time cost of updates, as in \S\ref{sec:core_results}.

We display results in Figure \ref{fig:sparse_upcycling}.   MoE substantially underperforms \cbtm with 16 clusters as the compute budget grows. Surprisingly, we observe that with enough compute, MoE underperforms even the dense LM, and that when compute budgets are further increased, losses consistently explode. This suggests that sparse upcycling is highly unstable, possibly due to distribution shifts from pretraining. 

In Figure \ref{fig:ups}, we compare the maximum seconds-per-update of the MoE model with that of the \cbtm models. MoE becomes substantially slower as more GPUs are used during training. This is largely due to expensive all-to-all communication that occurs between experts during MoE training, which is necessary to route tokens to experts \citep{artetxe2021efficient}. On the other hand, our method does not have any shared parameters between experts. Also, MoE expends more FLOPs during training than the \cbtm models. Finally, MoE still requires synchronous compute to train experts due to shared parameters, so they are also afflicted by the practical difficulties of training dense language models at scale \S\ref{sec:time_comparison}. 


\subsection{Summary}

Our results suggest that language models trained with \cbtm substantially outperform MoEs trained to the same budget. The performance gains of our technique likely are a result of the simplicity of our deterministic routing (based on empirically derived clusters), instabilities associated with sparse upcycling, and other factors.

\section{Analysis}
\label{sec:analysis}

In \S\ref{sec:results}, \S\ref{sec:downstream_tasks}, and \S\ref{sec:moe_comparison}, we demonstrate that \cbtm outperforms compute-matched densely trained and MoE baselines. We now study our clustering approach in more detail and describe its effect on overall performance of \cbtm.

\subsection{Is clustering important?}
\label{sec:analysis_random}

\begin{figure}[t!]
    \centering
    \includegraphics[width=\columnwidth]{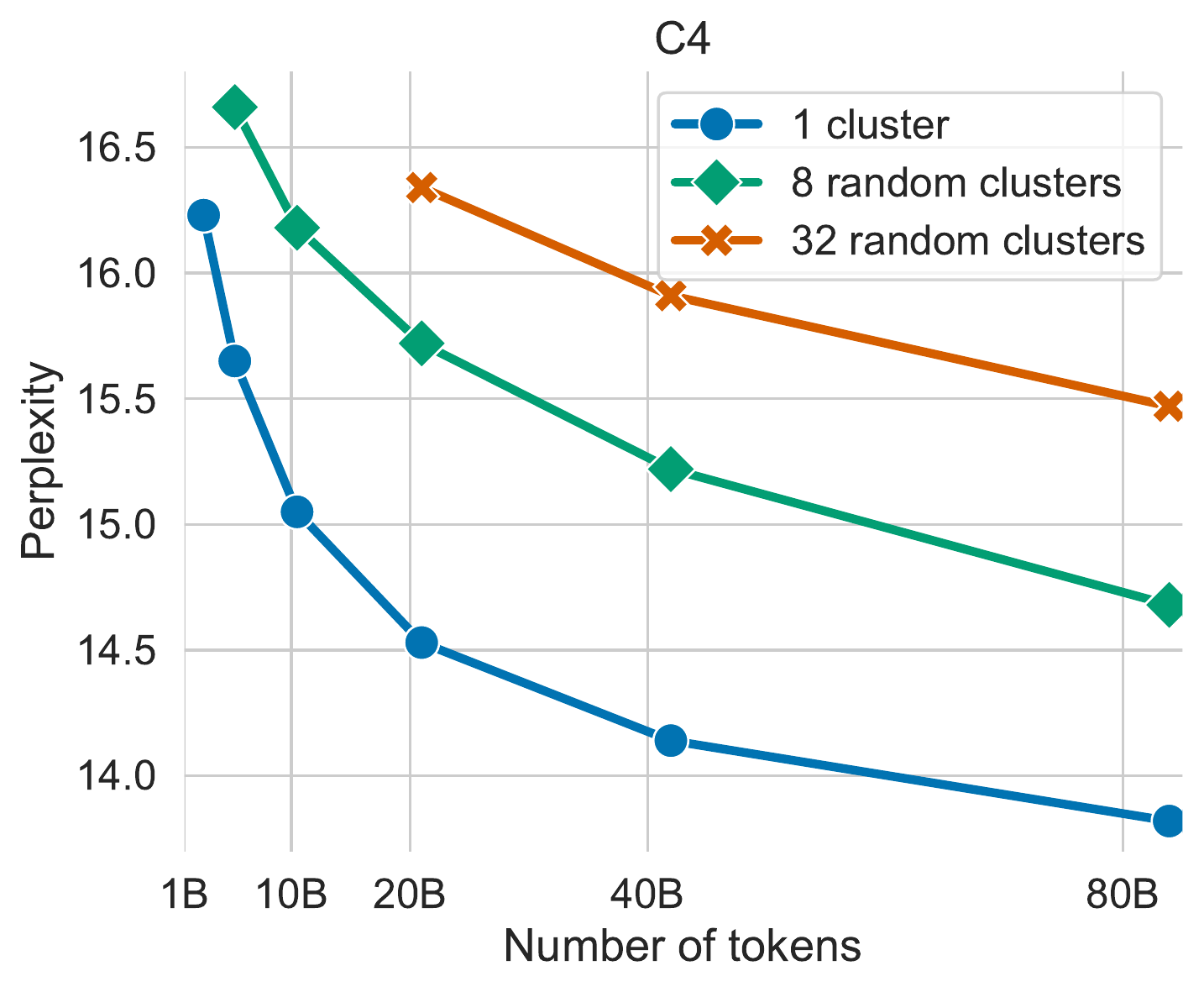} \vspace{-2em}
    \caption{\textbf{Random clusters underperform (\S\ref{sec:analysis_random}).} Training experts on random clusters underperforms even the dense, single cluster model, showing the importance of cluster specialization. Random clusters become more harmful as the cluster count grows.}
    \label{fig:random}
\end{figure}

To assess the importance of the clustering algorithm, we perform \cbtm as above, except that we assign each document to a random cluster, rather than a learned one. This is equivalent to the random ensemble baseline from \citet{btm}. Results in Figure~\ref{fig:random} demonstrate that using random clusters dramatically underperforms both our method and the dense baseline. Therefore, cluster specialization is vital for \cbtm. This confirms results from \citet{btm}, who found that domain specialization of ELMs is critical for performance the ensemble, as well as those from \citet{https://doi.org/10.48550/arxiv.2302.03202}, who show that instruction-specialized ELMs transfer to other tasks with similar instructions.

\subsection{Is it important to balance the clusters?}
\label{sec:analysis_balancing}

\begin{figure}[!t]
    \centering
    \includegraphics[width=\columnwidth]{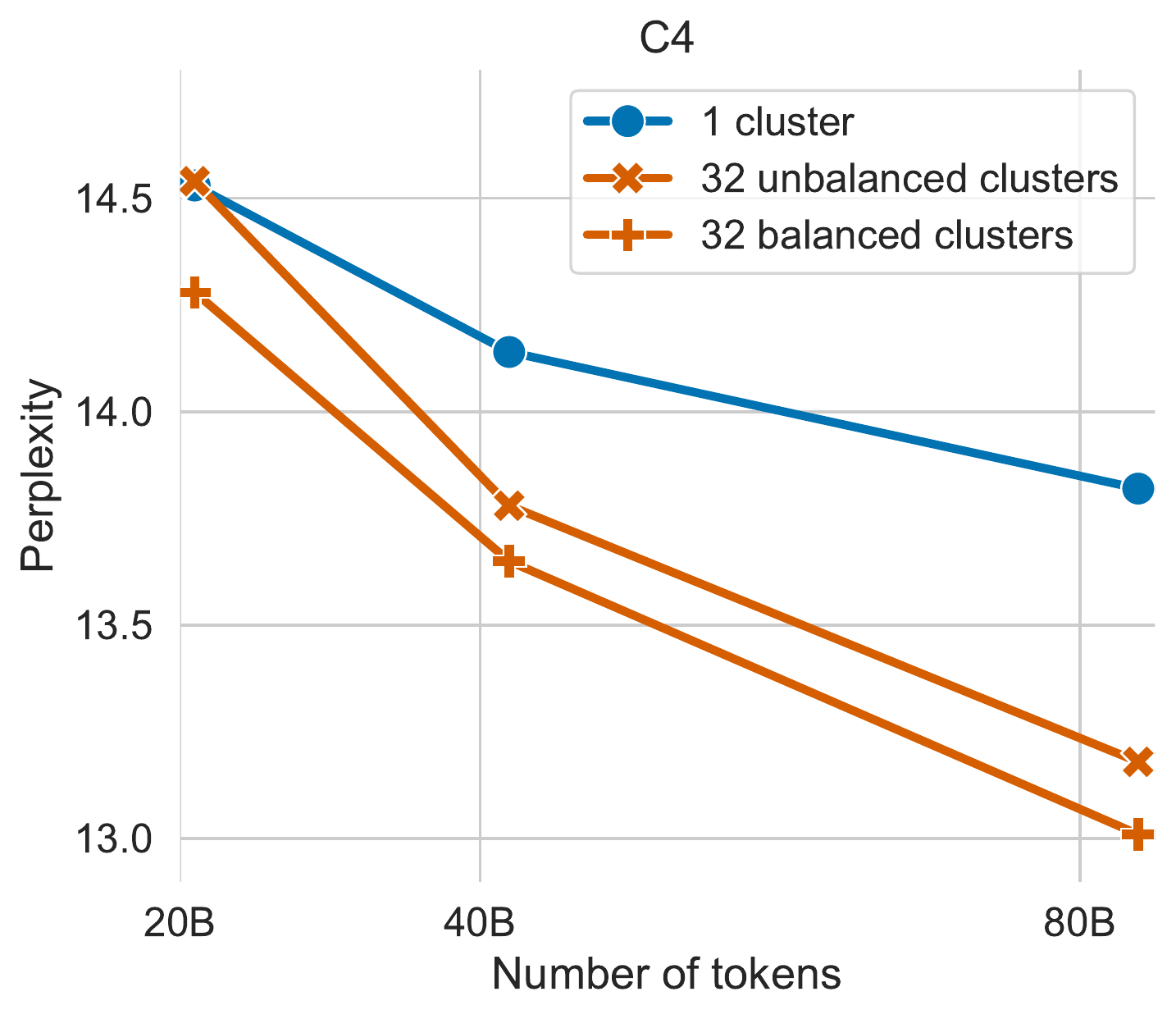} \vspace{-2em}
    \caption{\textbf{Cluster balancing improves performance (\S\ref{sec:analysis_balancing}).} When training on C4 with 32 clusters, balancing consistently improves over the unbalanced version, suggesting that cluster size important aspect to \cbtm. }
    \label{fig:unbalanced}
\end{figure}

Applying a balancing constraint in $k$-means  avoids the degenerate outcome of a long tail in cluster sizes (\citealt{https://doi.org/10.48550/arxiv.1411.6235}). Indeed, with 10K documents of held out validation data in C4, we observe that balanced clustering  significantly increases the median cluster size, and narrows its range, relative to an unbalanced baseline (\S\ref{sec:appendix_balancing}).  To assess the effect of balancing cluster size on the performance of \cbtm, we perform \cbtm with a $k$-means clustering model but remove the balancing constraint. For the 8-cluster model, we observe that balancing has little effect. However, for the 32-cluster model (Figure \ref{fig:unbalanced}), unbalanced clustering consistently leads to worse performance. These results suggest that balancing becomes more important as one scales the number of clusters. This is consistent with separate experiments that show that the long tail in cluster sizes becomes a more consistent problem with higher cluster counts.

\subsection{Are clusters well defined? Do experts specialize?}
\label{sec:analysis_specialze}

Since we use tf-idf as our document embedder in \cbtm, we can perform an inverse transform from the cluster centers into the vocabulary space to identify terms that most likely would have been embedded as the cluster center. We display the top five terms per cluster in \S\ref{sec:cluster_terms}. We observe that as the number of clusters increases, the top terms across clusters become more specific and varied.

\begin{figure}[t!]
    \centering
    \includegraphics[width=\columnwidth]{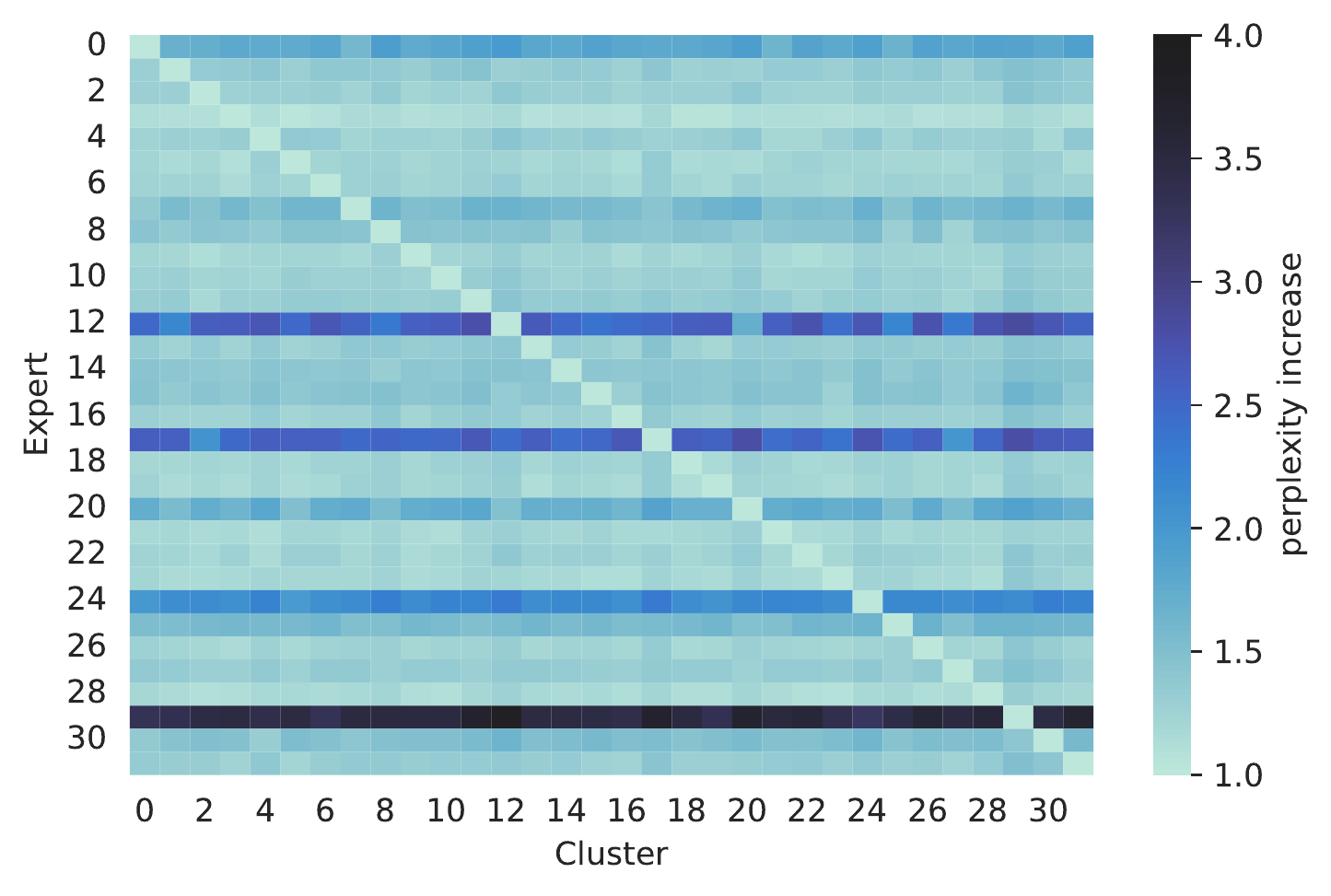} 
    \vspace{-2em}
    \caption{\textbf{Experts specialize to their cluster (\S\ref{sec:analysis_specialze}).} Here, we use the 32-cluster model trained on 84B tokens of C4. Each cell is a ratio between one expert’s test perplexity on a cluster to that of the expert trained on that cluster. The diagonal indicates that experts perform best on the cluster they were trained on. Many experts transfer well across clusters, but some do not.}
    \label{fig:heatmap}
\end{figure}

Next, we study whether ELMs trained on these clusters specialize. Using the 32-cluster model trained on 84B tokens of C4, we compute perplexity of all experts across 200 held out documents in each cluster. 
For each cluster, we then measure the ratio of the perplexity of each expert to the perplexity of the expert trained on that cluster. 
We display those ratios in Figure \ref{fig:heatmap}. We see that all experts perform best on their own cluster. Some experts do not transfer well at all to other clusters, while others do reasonably well. Cross referencing with the cluster term tables in  \S\ref{sec:cluster_terms}, we see that cluster experts 3 and 5 tend to generalize well and the top terms in these clusters are more generic (with words such as "\emph{just}, \emph{like}, \emph{love}").  The experts specialized to content such as \emph{"site, page, website"} (cluster 0) and \emph{"app, phone, video"} (cluster 29), tend to do poorly on all other clusters.\footnote{This result imply that cluster experts can be removed to filter out unwanted generations after training, without significantly impacting performance on other content. We leave such exploration to future work.} These results suggest that experts specialize to their cluster. We infer that the success of sparse \cbtm inference is a result of expert specialization, and that \cbtm performance gains may be partially due to the sample efficiency of specialized training.

\subsection{How do clusters and metadata domains compare?}
\label{sec:analysis_metadata}

\begin{figure}[t!]
    \centering
    \includegraphics[scale=0.5]{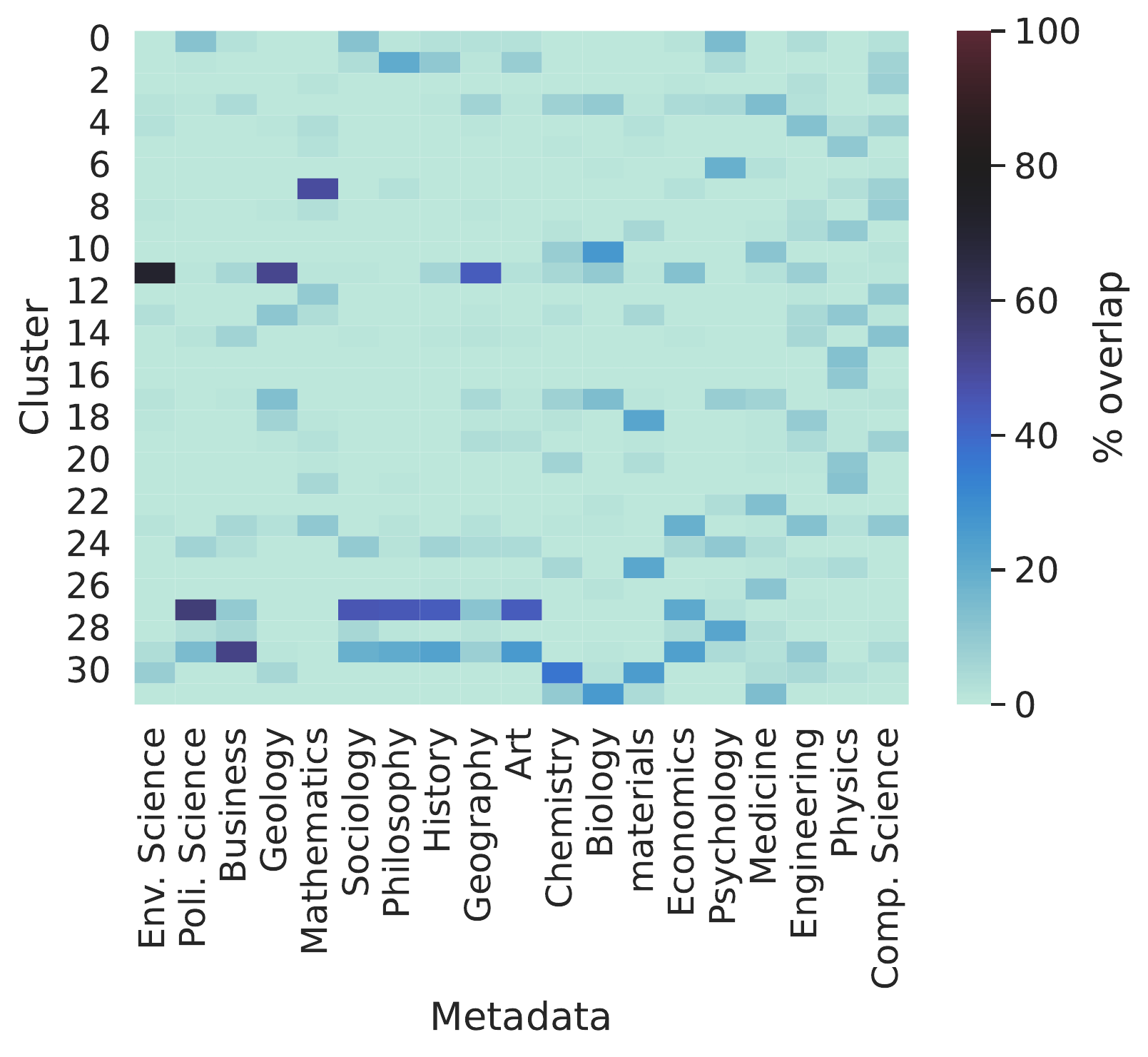}
    \caption{\textbf{Clusters and metadata do not perfectly align (\S\ref{sec:analysis_metadata}).} Each cell in the heatmap is the \% overlap between a cluster and  a metadata label identifying the field-of-study of a document in S2ORC; high overlap indicates that most documents with the corresponding label get assigned to the corresponding cluster. While documents with certain labels (e.g., Environmental Science, Political Science, Business) get primarily assigned to a single cluster, documents with other labels (e.g., Engineering, Physics, Computer Science) are distributed across multiple clusters.}
    \label{fig:s2orc_purity}
\end{figure}

The key motivation for \cbtm is to remove the reliance on metadata to delineate the domains to which ELMs specialize. How well do clusters reflect the dataset segmentation produced by metadata? We use S2ORC to study this question. First, we align the learned clusters from a 32-cluster model with the fields-of-study metadata available from S2ORC \citep{s2orc}. Then we visualize the overlap between the metadata and clusters (Figure \ref{fig:s2orc_purity}).   We observe only a partial alignment between metadata and clusters in S2ORC. Documents with some metadata labels (e.g., Enviromental Science, Political Science) are mostly assigned to their own clusters, while documents with other labels (e.g., Computer Science, Physics) are distributed across multiple clusters. 

The partial alignment between metadata and clusters suggests that \cbtm models may not have the same performance as those trained with metadata labels to delineate domains. To investigate this hypothesis further, we perform experiments using  a subset of the Pile \citep{pile} to compare the performance of experts trained with metadata and experts trained with clusters. See \S\ref{sec:metadata_comparison} for more details on this corpus. We observe that experts trained with learned clusters perform slightly better than those with metdata labels on a held out validation data (Table \ref{tab:metadatacompare} in the appendix). Both techniques perform better than training with just a single cluster on the Pile, confirming our results from \S\ref{sec:core_results}. These results imply that metadata may not correspond with the most optimal segmentation of the corpus. However, using metadata has the advantage of interpretability and simplicity, and metadata can identify domains that are not just lexically driven \citep[e.g.,][]{lucy-bamman-2021-characterizing, gururangan-etal-2022-demix}. Future work may explore combining metadata- and cluster-specialized ELMs.

\subsection{Summary} Our analysis demonstrates that the improvements from \cbtm are not the result of ensembling alone. Various components of our training method, particularly the nature of the learned clusters, play a critical role in \cbtm performance. Improving the representation of domains in a corpus, perhaps using other pretrained representations \citep{aharoni-goldberg-2020-unsupervised} or more sophisticated clustering algorithms \citep{10.5555/3001460.3001507,chronopoulou-etal-2022-efficient}, are likely to improve \cbtm performance.

\section{Related Work}

\paragraph{Sparse Models} \cbtm is closely related to sparse models which activate only a subset of  parameters \citep{pmlr-v119-evci20a,pmlr-v97-mostafa19a,dettmers-sparse-from-scratch}. \cbtm is inspired by MoE, but is much simpler and more efficient to train. Most MoE methods rely on training token-based routing mechanisms \citep{lepikhin2020gshard,fedus2021switch,baselayers,roller2021hash}, but others rely on task \citep{https://doi.org/10.48550/arxiv.2110.03742} or domain \citep{gururangan-etal-2022-demix} routing. 

\paragraph{Expert Language Models} As we note throughout the study, this work is most directly related to BTM \citep{btm}. BTM is in turn partially inspired by prior work on variations of MoE models \citep{jacobs1991adaptive}, but especially DEMix layers \citep{gururangan-etal-2022-demix}, which replace transformer feedforward layers with metadata-defined domain experts. \citet{https://doi.org/10.48550/arxiv.2302.03202} train expert language models on instruction-based tasks, while \citet{pfeiffer-etal-2022-lifting} train expert language models on different languages. 

\paragraph{Cluster Routing} \citet{chronopoulou-etal-2022-efficient} and \citet{Chronopoulou2023AdapterSoupWA} use hierarchical clustering to identify domains to specialize adapter experts to, and use the adapters in an ensemble or parameter average at inference time. \citet{Duan2021EnsLMEL} build ensembles of task-specific models by clustering the training data of supervised tasks. \citet{gross2017hard} employ a cluster-based router similar to ours in an image classification setting using ResNets. However, they use a hard routing (or only activate a single expert) in both training and inference, while we use hard routing during training but ensemble experts during inference. Our inference technique is inspired by nearest neighbor retrieval mechanisms in language models \citep{https://doi.org/10.48550/arxiv.1911.00172,https://doi.org/10.48550/arxiv.2205.13792}. 

\paragraph{Communication-efficient training} Our study contributes to a line of research into communication-efficient algorithms for training large models. Some previous work proposes ways to train large dense models collaboratively over distributed networks of servers \citep{https://doi.org/10.48550/arxiv.2209.01188, https://doi.org/10.48550/arxiv.2206.01288}. Other works focus on new forms of data \citep{https://doi.org/10.48550/arxiv.2107.01499}, model \citep{https://doi.org/10.48550/arxiv.2301.11913}, and pipeline \citep{https://doi.org/10.48550/arxiv.2206.01299} parallelism to allow for model training on heterogeneous devices that can recover from node failures. \citet{lofi} propose a communication-efficient method of fine-tuning by training a collection of models with different hyperparameters on individual GPU nodes, and then averaging their parameters after training. Our work uses expert specialization for communication efficient training, and \cbtm can be combined with any of these other techniques to improve training efficiency.


\section{Conclusion}

We introduce \cbtm, a new technique to efficiently train sparse LMs. \cbtm splits a corpus into $k$ clusters, trains an expert LM on each cluster, and creates a sparse ensemble during inference. We observe that the optimal number of clusters for \cbtm increases with the amount of data, and using more clusters also allows us to aggressively parallelize training to efficiently scale into massive datasets. Future work could investigate \cbtm in multitask or multilingual settings, the usefulness of multiple iterations of \cbtm on a corpus (perhaps with hierarchical clustering), or the possibility of combining metadata- and cluster-based routing to scale into many heterogeneous datasets in parallel. 

\section*{Acknowledgements}

This paper benefited from thoughtful feedback from a number of people: Armen Aghajanyan,  Tim Dettmers, Sneha Kudugunta, Stephen Roller, Swabha Swayamdipta, and Mitchell Wortsman.
\newpage
\bibliography{custom}

\begin{thebibliography}{60}
\providecommand{\natexlab}[1]{#1}
\providecommand{\url}[1]{\texttt{#1}}
\expandafter\ifx\csname urlstyle\endcsname\relax
  \providecommand{\doi}[1]{doi: #1}\else
  \providecommand{\doi}{doi: \begingroup \urlstyle{rm}\Url}\fi

\bibitem[Aharoni \& Goldberg(2020)Aharoni and
  Goldberg]{aharoni-goldberg-2020-unsupervised}
Aharoni, R. and Goldberg, Y.
\newblock Unsupervised domain clusters in pretrained language models.
\newblock In \emph{Proceedings of the 58th Annual Meeting of the Association
  for Computational Linguistics}, pp.\  7747--7763, Online, July 2020.
  Association for Computational Linguistics.
\newblock \doi{10.18653/v1/2020.acl-main.692}.
\newblock URL \url{https://aclanthology.org/2020.acl-main.692}.

\bibitem[Artetxe et~al.(2021)Artetxe, Bhosale, Goyal, Mihaylov, Ott, Shleifer,
  Lin, Du, Iyer, Pasunuru, Anantharaman, Li, Chen, Akin, Baines, Martin, Zhou,
  Koura, O'Horo, Wang, Zettlemoyer, Diab, Kozareva, and
  Stoyanov]{artetxe2021efficient}
Artetxe, M., Bhosale, S., Goyal, N., Mihaylov, T., Ott, M., Shleifer, S., Lin,
  X.~V., Du, J., Iyer, S., Pasunuru, R., Anantharaman, G., Li, X., Chen, S.,
  Akin, H., Baines, M., Martin, L., Zhou, X., Koura, P.~S., O'Horo, B., Wang,
  J., Zettlemoyer, L., Diab, M., Kozareva, Z., and Stoyanov, V.
\newblock Efficient large scale language modeling with mixtures of experts,
  2021.
\newblock URL \url{https://arxiv.org/abs/2112.10684}.

\bibitem[Arthur \& Vassilvitskii(2007)Arthur and
  Vassilvitskii]{10.5555/1283383.1283494}
Arthur, D. and Vassilvitskii, S.
\newblock K-means++: The advantages of careful seeding.
\newblock In \emph{Proceedings of the Eighteenth Annual ACM-SIAM Symposium on
  Discrete Algorithms}, SODA '07, pp.\  1027–1035, USA, 2007. Society for
  Industrial and Applied Mathematics.
\newblock ISBN 9780898716245.

\bibitem[Barbieri et~al.(2020)Barbieri, Camacho-Collados, Espinosa~Anke, and
  Neves]{barbieri-etal-2020-tweeteval}
Barbieri, F., Camacho-Collados, J., Espinosa~Anke, L., and Neves, L.
\newblock {T}weet{E}val: Unified benchmark and comparative evaluation for tweet
  classification.
\newblock In \emph{Findings of the Association for Computational Linguistics:
  EMNLP 2020}, pp.\  1644--1650, Online, November 2020. Association for
  Computational Linguistics.
\newblock \doi{10.18653/v1/2020.findings-emnlp.148}.
\newblock URL \url{https://aclanthology.org/2020.findings-emnlp.148}.

\bibitem[Bertsekas(1992)]{Bertsekas1992AuctionAF}
Bertsekas, D.~P.
\newblock Auction algorithms for network flow problems: A tutorial
  introduction.
\newblock \emph{Computational Optimization and Applications}, 1:\penalty0
  7--66, 1992.

\bibitem[Borzunov et~al.(2022)Borzunov, Baranchuk, Dettmers, Ryabinin, Belkada,
  Chumachenko, Samygin, and Raffel]{https://doi.org/10.48550/arxiv.2209.01188}
Borzunov, A., Baranchuk, D., Dettmers, T., Ryabinin, M., Belkada, Y.,
  Chumachenko, A., Samygin, P., and Raffel, C.
\newblock Petals: Collaborative inference and fine-tuning of large models,
  2022.
\newblock URL \url{https://arxiv.org/abs/2209.01188}.

\bibitem[Chang et~al.(2014)Chang, Nie, Ma, and
  Yang]{https://doi.org/10.48550/arxiv.1411.6235}
Chang, X., Nie, F., Ma, Z., and Yang, Y.
\newblock Balanced k-means and min-cut clustering, 2014.
\newblock URL \url{https://arxiv.org/abs/1411.6235}.

\bibitem[Chowdhery et~al.(2022)Chowdhery, Narang, Devlin, Bosma, Mishra,
  Roberts, Barham, Chung, Sutton, Gehrmann, Schuh, Shi, Tsvyashchenko, Maynez,
  Rao, Barnes, Tay, Shazeer, Prabhakaran, Reif, Du, Hutchinson, Pope, Bradbury,
  Austin, Isard, Gur-Ari, Yin, Duke, Levskaya, Ghemawat, Dev, Michalewski,
  Garcia, Misra, Robinson, Fedus, Zhou, Ippolito, Luan, Lim, Zoph, Spiridonov,
  Sepassi, Dohan, Agrawal, Omernick, Dai, Pillai, Pellat, Lewkowycz, Moreira,
  Child, Polozov, Lee, Zhou, Wang, Saeta, Diaz, Firat, Catasta, Wei,
  Meier-Hellstern, Eck, Dean, Petrov, and Fiedel]{palm}
Chowdhery, A., Narang, S., Devlin, J., Bosma, M., Mishra, G., Roberts, A.,
  Barham, P., Chung, H.~W., Sutton, C., Gehrmann, S., Schuh, P., Shi, K.,
  Tsvyashchenko, S., Maynez, J., Rao, A., Barnes, P., Tay, Y., Shazeer, N.,
  Prabhakaran, V., Reif, E., Du, N., Hutchinson, B., Pope, R., Bradbury, J.,
  Austin, J., Isard, M., Gur-Ari, G., Yin, P., Duke, T., Levskaya, A.,
  Ghemawat, S., Dev, S., Michalewski, H., Garcia, X., Misra, V., Robinson, K.,
  Fedus, L., Zhou, D., Ippolito, D., Luan, D., Lim, H., Zoph, B., Spiridonov,
  A., Sepassi, R., Dohan, D., Agrawal, S., Omernick, M., Dai, A.~M., Pillai,
  T.~S., Pellat, M., Lewkowycz, A., Moreira, E., Child, R., Polozov, O., Lee,
  K., Zhou, Z., Wang, X., Saeta, B., Diaz, M., Firat, O., Catasta, M., Wei, J.,
  Meier-Hellstern, K., Eck, D., Dean, J., Petrov, S., and Fiedel, N.
\newblock Palm: Scaling language modeling with pathways, 2022.
\newblock URL \url{https://arxiv.org/abs/2204.02311}.

\bibitem[Chronopoulou et~al.(2022)Chronopoulou, Peters, and
  Dodge]{chronopoulou-etal-2022-efficient}
Chronopoulou, A., Peters, M., and Dodge, J.
\newblock Efficient hierarchical domain adaptation for pretrained language
  models.
\newblock In \emph{Proceedings of the 2022 Conference of the North American
  Chapter of the Association for Computational Linguistics: Human Language
  Technologies}, pp.\  1336--1351, Seattle, United States, July 2022.
  Association for Computational Linguistics.
\newblock \doi{10.18653/v1/2022.naacl-main.96}.
\newblock URL \url{https://aclanthology.org/2022.naacl-main.96}.

\bibitem[Chronopoulou et~al.(2023)Chronopoulou, Peters, Fraser, and
  Dodge]{Chronopoulou2023AdapterSoupWA}
Chronopoulou, A., Peters, M.~E., Fraser, A.~M., and Dodge, J.
\newblock Adaptersoup: Weight averaging to improve generalization of pretrained
  language models.
\newblock \emph{ArXiv}, abs/2302.07027, 2023.

\bibitem[Clark et~al.(2022)Clark, Casas, Guy, Mensch, Paganini, Hoffmann,
  Damoc, Hechtman, Cai, Borgeaud, Driessche, Rutherford, Hennigan, Johnson,
  Millican, Cassirer, Jones, Buchatskaya, Budden, Sifre, Osindero, Vinyals,
  Rae, Elsen, Kavukcuoglu, and
  Simonyan]{https://doi.org/10.48550/arxiv.2202.01169}
Clark, A., Casas, D. d.~l., Guy, A., Mensch, A., Paganini, M., Hoffmann, J.,
  Damoc, B., Hechtman, B., Cai, T., Borgeaud, S., Driessche, G. v.~d.,
  Rutherford, E., Hennigan, T., Johnson, M., Millican, K., Cassirer, A., Jones,
  C., Buchatskaya, E., Budden, D., Sifre, L., Osindero, S., Vinyals, O., Rae,
  J., Elsen, E., Kavukcuoglu, K., and Simonyan, K.
\newblock Unified scaling laws for routed language models, 2022.
\newblock URL \url{https://arxiv.org/abs/2202.01169}.

\bibitem[Dehghani et~al.(2021)Dehghani, Arnab, Beyer, Vaswani, and
  Tay]{https://doi.org/10.48550/arxiv.2110.12894}
Dehghani, M., Arnab, A., Beyer, L., Vaswani, A., and Tay, Y.
\newblock The efficiency misnomer, 2021.
\newblock URL \url{https://arxiv.org/abs/2110.12894}.

\bibitem[Dettmers \& Zettlemoyer(2019)Dettmers and
  Zettlemoyer]{dettmers-sparse-from-scratch}
Dettmers, T. and Zettlemoyer, L.
\newblock Sparse networks from scratch: Faster training without losing
  performance.
\newblock \emph{CoRR}, abs/1907.04840, 2019.
\newblock URL \url{http://arxiv.org/abs/1907.04840}.

\bibitem[Duan et~al.(2021)Duan, Zhang, Wang, Wang, Chen, and
  Zhou]{Duan2021EnsLMEL}
Duan, Z., Zhang, H., Wang, C., Wang, Z., Chen, B., and Zhou, M.
\newblock Enslm: Ensemble language model for data diversity by semantic
  clustering.
\newblock In \emph{Annual Meeting of the Association for Computational
  Linguistics}, 2021.

\bibitem[Ester et~al.(1996)Ester, Kriegel, Sander, and
  Xu]{10.5555/3001460.3001507}
Ester, M., Kriegel, H.-P., Sander, J., and Xu, X.
\newblock A density-based algorithm for discovering clusters in large spatial
  databases with noise.
\newblock In \emph{Proceedings of the Second International Conference on
  Knowledge Discovery and Data Mining}, KDD'96, pp.\  226–231. AAAI Press,
  1996.

\bibitem[Evci et~al.(2020)Evci, Gale, Menick, Castro, and
  Elsen]{pmlr-v119-evci20a}
Evci, U., Gale, T., Menick, J., Castro, P.~S., and Elsen, E.
\newblock Rigging the lottery: Making all tickets winners.
\newblock In III, H.~D. and Singh, A. (eds.), \emph{Proceedings of the 37th
  International Conference on Machine Learning}, volume 119 of
  \emph{Proceedings of Machine Learning Research}, pp.\  2943--2952. PMLR,
  13--18 Jul 2020.
\newblock URL \url{https://proceedings.mlr.press/v119/evci20a.html}.

\bibitem[Fedus et~al.(2021)Fedus, Zoph, and Shazeer]{fedus2021switch}
Fedus, W., Zoph, B., and Shazeer, N.
\newblock Switch transformers: Scaling to trillion parameter models with simple
  and efficient sparsity.
\newblock \emph{J. Mach. Learn. Res}, 23:\penalty0 1--40, 2021.

\bibitem[Fedus et~al.(2022)Fedus, Dean, and
  Zoph]{https://doi.org/10.48550/arxiv.2209.01667}
Fedus, W., Dean, J., and Zoph, B.
\newblock A review of sparse expert models in deep learning, 2022.
\newblock URL \url{https://arxiv.org/abs/2209.01667}.

\bibitem[Gan et~al.(2021)Gan, Lian, Wang, Chang, Liu, Shi, Zhang, Li, Sun,
  Jiang, Yuan, Yang, Liu, and Zhang]{https://doi.org/10.48550/arxiv.2107.01499}
Gan, S., Lian, X., Wang, R., Chang, J., Liu, C., Shi, H., Zhang, S., Li, X.,
  Sun, T., Jiang, J., Yuan, B., Yang, S., Liu, J., and Zhang, C.
\newblock Bagua: Scaling up distributed learning with system relaxations, 2021.
\newblock URL \url{https://arxiv.org/abs/2107.01499}.

\bibitem[Gao et~al.(2021)Gao, Biderman, Black, Golding, Hoppe, Foster, Phang,
  He, Thite, Nabeshima, Presser, and Leahy]{pile}
Gao, L., Biderman, S., Black, S., Golding, L., Hoppe, T., Foster, C., Phang,
  J., He, H., Thite, A., Nabeshima, N., Presser, S., and Leahy, C.
\newblock The pile: An 800gb dataset of diverse text for language modeling,
  2021.
\newblock URL \url{https://arxiv.org/abs/2101.00027}.

\bibitem[Gross et~al.(2017)Gross, Ranzato, and Szlam]{gross2017hard}
Gross, S., Ranzato, M., and Szlam, A.
\newblock Hard mixtures of experts for large scale weakly supervised vision.
\newblock In \emph{Proceedings of the IEEE Conference on Computer Vision and
  Pattern Recognition}, pp.\  6865--6873, 2017.

\bibitem[Gururangan et~al.(2022)Gururangan, Lewis, Holtzman, Smith, and
  Zettlemoyer]{gururangan-etal-2022-demix}
Gururangan, S., Lewis, M., Holtzman, A., Smith, N.~A., and Zettlemoyer, L.
\newblock {DEM}ix layers: Disentangling domains for modular language modeling.
\newblock In \emph{Proceedings of the 2022 Conference of the North American
  Chapter of the Association for Computational Linguistics: Human Language
  Technologies}, pp.\  5557--5576, Seattle, United States, July 2022.
  Association for Computational Linguistics.
\newblock \doi{10.18653/v1/2022.naacl-main.407}.
\newblock URL \url{https://aclanthology.org/2022.naacl-main.407}.

\bibitem[Hoffmann et~al.(2022)Hoffmann, Borgeaud, Mensch, Buchatskaya, Cai,
  Rutherford, de~Las~Casas, Hendricks, Welbl, Clark,
  et~al.]{hoffmann2022training}
Hoffmann, J., Borgeaud, S., Mensch, A., Buchatskaya, E., Cai, T., Rutherford,
  E., de~Las~Casas, D., Hendricks, L.~A., Welbl, J., Clark, A., et~al.
\newblock Training compute-optimal large language models.
\newblock 2022.

\bibitem[Jacobs et~al.(1991)Jacobs, Jordan, Nowlan, and
  Hinton]{jacobs1991adaptive}
Jacobs, R.~A., Jordan, M.~I., Nowlan, S.~J., and Hinton, G.~E.
\newblock Adaptive mixtures of local experts.
\newblock \emph{Neural computation}, 3\penalty0 (1):\penalty0 79--87, 1991.

\bibitem[Jang et~al.(2023)Jang, Kim, Ye, Kim, Logeswaran, Lee, Lee, and
  Seo]{https://doi.org/10.48550/arxiv.2302.03202}
Jang, J., Kim, S., Ye, S., Kim, D., Logeswaran, L., Lee, M., Lee, K., and Seo,
  M.
\newblock Exploring the benefits of training expert language models over
  instruction tuning, 2023.
\newblock URL \url{https://arxiv.org/abs/2302.03202}.

\bibitem[Khandelwal et~al.(2019)Khandelwal, Levy, Jurafsky, Zettlemoyer, and
  Lewis]{https://doi.org/10.48550/arxiv.1911.00172}
Khandelwal, U., Levy, O., Jurafsky, D., Zettlemoyer, L., and Lewis, M.
\newblock Generalization through memorization: Nearest neighbor language
  models, 2019.
\newblock URL \url{https://arxiv.org/abs/1911.00172}.

\bibitem[Komatsuzaki et~al.(2022)Komatsuzaki, Puigcerver, Lee-Thorp, Ruiz,
  Mustafa, Ainslie, Tay, Dehghani, and
  Houlsby]{https://doi.org/10.48550/arxiv.2212.05055}
Komatsuzaki, A., Puigcerver, J., Lee-Thorp, J., Ruiz, C.~R., Mustafa, B.,
  Ainslie, J., Tay, Y., Dehghani, M., and Houlsby, N.
\newblock Sparse upcycling: Training mixture-of-experts from dense checkpoints,
  2022.
\newblock URL \url{https://arxiv.org/abs/2212.05055}.

\bibitem[Kudugunta et~al.(2021)Kudugunta, Huang, Bapna, Krikun, Lepikhin,
  Luong, and Firat]{https://doi.org/10.48550/arxiv.2110.03742}
Kudugunta, S., Huang, Y., Bapna, A., Krikun, M., Lepikhin, D., Luong, M.-T.,
  and Firat, O.
\newblock Beyond distillation: Task-level mixture-of-experts for efficient
  inference, 2021.
\newblock URL \url{https://arxiv.org/abs/2110.03742}.

\bibitem[Lehmann et~al.(2014)Lehmann, Isele, Jakob, Jentzsch, Kontokostas,
  Mendes, Hellmann, Morsey, Van~Kleef, Auer, and Bizer]{dbpedia}
Lehmann, J., Isele, R., Jakob, M., Jentzsch, A., Kontokostas, D., Mendes, P.,
  Hellmann, S., Morsey, M., Van~Kleef, P., Auer, S., and Bizer, C.
\newblock Dbpedia - a large-scale, multilingual knowledge base extracted from
  wikipedia.
\newblock \emph{Semantic Web Journal}, 6, 01 2014.
\newblock \doi{10.3233/SW-140134}.

\bibitem[Lepikhin et~al.(2021)Lepikhin, Lee, Xu, Chen, Firat, Huang, Krikun,
  Shazeer, and Chen]{lepikhin2020gshard}
Lepikhin, D., Lee, H., Xu, Y., Chen, D., Firat, O., Huang, Y., Krikun, M.,
  Shazeer, N., and Chen, Z.
\newblock {\{}GS{\}}hard: Scaling giant models with conditional computation and
  automatic sharding.
\newblock In \emph{International Conference on Learning Representations}, 2021.
\newblock URL \url{https://openreview.net/forum?id=qrwe7XHTmYb}.

\bibitem[Lewis et~al.(2021)Lewis, Bhosale, Dettmers, Goyal, and
  Zettlemoyer]{baselayers}
Lewis, M., Bhosale, S., Dettmers, T., Goyal, N., and Zettlemoyer, L.
\newblock Base layers: Simplifying training of large, sparse models, 2021.
\newblock URL \url{https://arxiv.org/abs/2103.16716}.

\bibitem[Li et~al.(2022)Li, Gururangan, Dettmers, Lewis, Althoff, Smith, and
  Zettlemoyer]{btm}
Li, M., Gururangan, S., Dettmers, T., Lewis, M., Althoff, T., Smith, N.~A., and
  Zettlemoyer, L.
\newblock Branch-train-merge: Embarrassingly parallel training of expert
  language models, 2022.
\newblock URL \url{https://arxiv.org/abs/2208.03306}.

\bibitem[Lo et~al.(2019)Lo, Wang, Neumann, Kinney, and Weld]{s2orc}
Lo, K., Wang, L.~L., Neumann, M., Kinney, R., and Weld, D.~S.
\newblock S2orc: The semantic scholar open research corpus, 2019.
\newblock URL \url{https://arxiv.org/abs/1911.02782}.

\bibitem[Lu et~al.(2022)Lu, Bartolo, Moore, Riedel, and
  Stenetorp]{lu-etal-2022-fantastically}
Lu, Y., Bartolo, M., Moore, A., Riedel, S., and Stenetorp, P.
\newblock Fantastically ordered prompts and where to find them: Overcoming
  few-shot prompt order sensitivity.
\newblock In \emph{Proceedings of the 60th Annual Meeting of the Association
  for Computational Linguistics (Volume 1: Long Papers)}, pp.\  8086--8098,
  Dublin, Ireland, May 2022. Association for Computational Linguistics.
\newblock \doi{10.18653/v1/2022.acl-long.556}.
\newblock URL \url{https://aclanthology.org/2022.acl-long.556}.

\bibitem[Lucy \& Bamman(2021)Lucy and Bamman]{lucy-bamman-2021-characterizing}
Lucy, L. and Bamman, D.
\newblock Characterizing {E}nglish variation across social media communities
  with {BERT}.
\newblock \emph{Transactions of the Association for Computational Linguistics},
  9:\penalty0 538--556, 2021.
\newblock \doi{10.1162/tacl_a_00383}.
\newblock URL \url{https://aclanthology.org/2021.tacl-1.33}.

\bibitem[Maas et~al.(2011)Maas, Daly, Pham, Huang, Ng, and
  Potts]{maas-etal-2011-learning}
Maas, A.~L., Daly, R.~E., Pham, P.~T., Huang, D., Ng, A.~Y., and Potts, C.
\newblock Learning word vectors for sentiment analysis.
\newblock In \emph{Proceedings of the 49th Annual Meeting of the Association
  for Computational Linguistics: Human Language Technologies}, pp.\  142--150,
  Portland, Oregon, USA, June 2011. Association for Computational Linguistics.
\newblock URL \url{https://aclanthology.org/P11-1015}.

\bibitem[Malinen \& Fr{\"a}nti(2014)Malinen and
  Fr{\"a}nti]{Malinen2014BalancedKF}
Malinen, M.~I. and Fr{\"a}nti, P.
\newblock Balanced k-means for clustering.
\newblock In \emph{International Workshop on Structural and Syntactic Pattern
  Recognition}, 2014.

\bibitem[Malo et~al.(2014)Malo, Sinha, Korhonen, Wallenius, and
  Takala]{Malo2014GoodDO}
Malo, P., Sinha, A., Korhonen, P., Wallenius, J., and Takala, P.
\newblock Good debt or bad debt: Detecting semantic orientations in economic
  texts.
\newblock \emph{Journal of the Association for Information Science and
  Technology}, 65, 2014.

\bibitem[McCandlish et~al.(2018)McCandlish, Kaplan, Amodei, and
  Team]{https://doi.org/10.48550/arxiv.1812.06162}
McCandlish, S., Kaplan, J., Amodei, D., and Team, O.~D.
\newblock An empirical model of large-batch training, 2018.
\newblock URL \url{https://arxiv.org/abs/1812.06162}.

\bibitem[Min et~al.(2022)Min, Lyu, Holtzman, Artetxe, Lewis, Hajishirzi, and
  Zettlemoyer]{min-etal-2022-rethinking}
Min, S., Lyu, X., Holtzman, A., Artetxe, M., Lewis, M., Hajishirzi, H., and
  Zettlemoyer, L.
\newblock Rethinking the role of demonstrations: What makes in-context learning
  work?
\newblock In \emph{Proceedings of the 2022 Conference on Empirical Methods in
  Natural Language Processing}, pp.\  11048--11064, Abu Dhabi, United Arab
  Emirates, December 2022. Association for Computational Linguistics.
\newblock URL \url{https://aclanthology.org/2022.emnlp-main.759}.

\bibitem[Mostafa \& Wang(2019)Mostafa and Wang]{pmlr-v97-mostafa19a}
Mostafa, H. and Wang, X.
\newblock Parameter efficient training of deep convolutional neural networks by
  dynamic sparse reparameterization.
\newblock In Chaudhuri, K. and Salakhutdinov, R. (eds.), \emph{Proceedings of
  the 36th International Conference on Machine Learning}, volume~97 of
  \emph{Proceedings of Machine Learning Research}, pp.\  4646--4655. PMLR,
  09--15 Jun 2019.
\newblock URL \url{https://proceedings.mlr.press/v97/mostafa19a.html}.

\bibitem[Ott et~al.(2019)Ott, Edunov, Baevski, Fan, Gross, Ng, Grangier, and
  Auli]{https://doi.org/10.48550/arxiv.1904.01038}
Ott, M., Edunov, S., Baevski, A., Fan, A., Gross, S., Ng, N., Grangier, D., and
  Auli, M.
\newblock fairseq: A fast, extensible toolkit for sequence modeling, 2019.
\newblock URL \url{https://arxiv.org/abs/1904.01038}.

\bibitem[Pfeiffer et~al.(2022)Pfeiffer, Goyal, Lin, Li, Cross, Riedel, and
  Artetxe]{pfeiffer-etal-2022-lifting}
Pfeiffer, J., Goyal, N., Lin, X., Li, X., Cross, J., Riedel, S., and Artetxe,
  M.
\newblock Lifting the curse of multilinguality by pre-training modular
  transformers.
\newblock In \emph{Proceedings of the 2022 Conference of the North American
  Chapter of the Association for Computational Linguistics: Human Language
  Technologies}, pp.\  3479--3495, Seattle, United States, July 2022.
  Association for Computational Linguistics.
\newblock \doi{10.18653/v1/2022.naacl-main.255}.
\newblock URL \url{https://aclanthology.org/2022.naacl-main.255}.

\bibitem[Radford et~al.(2019)Radford, Wu, Child, Luan, Amodei, and
  Sutskever]{radfordlanguage}
Radford, A., Wu, J., Child, R., Luan, D., Amodei, D., and Sutskever, I.
\newblock Language models are unsupervised multitask learners.
\newblock 2019.

\bibitem[Rae et~al.(2021)Rae, Borgeaud, Cai, Millican, Hoffmann, Song,
  Aslanides, Henderson, Ring, Young, Rutherford, Hennigan, Menick, Cassirer,
  Powell, Driessche, Hendricks, Rauh, Huang, Glaese, Welbl, Dathathri, Huang,
  Uesato, Mellor, Higgins, Creswell, McAleese, Wu, Elsen, Jayakumar,
  Buchatskaya, Budden, Sutherland, Simonyan, Paganini, Sifre, Martens, Li,
  Kuncoro, Nematzadeh, Gribovskaya, Donato, Lazaridou, Mensch, Lespiau,
  Tsimpoukelli, Grigorev, Fritz, Sottiaux, Pajarskas, Pohlen, Gong, Toyama,
  d'Autume, Li, Terzi, Mikulik, Babuschkin, Clark, Casas, Guy, Jones, Bradbury,
  Johnson, Hechtman, Weidinger, Gabriel, Isaac, Lockhart, Osindero, Rimell,
  Dyer, Vinyals, Ayoub, Stanway, Bennett, Hassabis, Kavukcuoglu, and
  Irving]{gopher}
Rae, J.~W., Borgeaud, S., Cai, T., Millican, K., Hoffmann, J., Song, F.,
  Aslanides, J., Henderson, S., Ring, R., Young, S., Rutherford, E., Hennigan,
  T., Menick, J., Cassirer, A., Powell, R., Driessche, G. v.~d., Hendricks,
  L.~A., Rauh, M., Huang, P.-S., Glaese, A., Welbl, J., Dathathri, S., Huang,
  S., Uesato, J., Mellor, J., Higgins, I., Creswell, A., McAleese, N., Wu, A.,
  Elsen, E., Jayakumar, S., Buchatskaya, E., Budden, D., Sutherland, E.,
  Simonyan, K., Paganini, M., Sifre, L., Martens, L., Li, X.~L., Kuncoro, A.,
  Nematzadeh, A., Gribovskaya, E., Donato, D., Lazaridou, A., Mensch, A.,
  Lespiau, J.-B., Tsimpoukelli, M., Grigorev, N., Fritz, D., Sottiaux, T.,
  Pajarskas, M., Pohlen, T., Gong, Z., Toyama, D., d'Autume, C. d.~M., Li, Y.,
  Terzi, T., Mikulik, V., Babuschkin, I., Clark, A., Casas, D. d.~L., Guy, A.,
  Jones, C., Bradbury, J., Johnson, M., Hechtman, B., Weidinger, L., Gabriel,
  I., Isaac, W., Lockhart, E., Osindero, S., Rimell, L., Dyer, C., Vinyals, O.,
  Ayoub, K., Stanway, J., Bennett, L., Hassabis, D., Kavukcuoglu, K., and
  Irving, G.
\newblock Scaling language models: Methods, analysis \& insights from training
  gopher, 2021.
\newblock URL \url{https://arxiv.org/abs/2112.11446}.

\bibitem[Raffel et~al.(2019)Raffel, Shazeer, Roberts, Lee, Narang, Matena,
  Zhou, Li, and Liu]{t5}
Raffel, C., Shazeer, N., Roberts, A., Lee, K., Narang, S., Matena, M., Zhou,
  Y., Li, W., and Liu, P.~J.
\newblock Exploring the limits of transfer learning with a unified text-to-text
  transformer, 2019.
\newblock URL \url{https://arxiv.org/abs/1910.10683}.

\bibitem[Roller et~al.(2021)Roller, Sukhbaatar, Szlam, and
  Weston]{roller2021hash}
Roller, S., Sukhbaatar, S., Szlam, A., and Weston, J.~E.
\newblock Hash layers for large sparse models.
\newblock In Beygelzimer, A., Dauphin, Y., Liang, P., and Vaughan, J.~W.
  (eds.), \emph{Advances in Neural Information Processing Systems}, 2021.
\newblock URL \url{https://openreview.net/forum?id=lMgDDWb1ULW}.

\bibitem[Ryabinin et~al.(2023)Ryabinin, Dettmers, Diskin, and
  Borzunov]{https://doi.org/10.48550/arxiv.2301.11913}
Ryabinin, M., Dettmers, T., Diskin, M., and Borzunov, A.
\newblock Swarm parallelism: Training large models can be surprisingly
  communication-efficient, 2023.
\newblock URL \url{https://arxiv.org/abs/2301.11913}.

\bibitem[Shallue et~al.(2018)Shallue, Lee, Antognini, Sohl-Dickstein, Frostig,
  and Dahl]{https://doi.org/10.48550/arxiv.1811.03600}
Shallue, C.~J., Lee, J., Antognini, J., Sohl-Dickstein, J., Frostig, R., and
  Dahl, G.~E.
\newblock Measuring the effects of data parallelism on neural network training.
\newblock 2018.
\newblock \doi{10.48550/ARXIV.1811.03600}.
\newblock URL \url{https://arxiv.org/abs/1811.03600}.

\bibitem[Shi et~al.(2022)Shi, Michael, Gururangan, and
  Zettlemoyer]{https://doi.org/10.48550/arxiv.2205.13792}
Shi, W., Michael, J., Gururangan, S., and Zettlemoyer, L.
\newblock knn-prompt: Nearest neighbor zero-shot inference, 2022.
\newblock URL \url{https://arxiv.org/abs/2205.13792}.

\bibitem[Svenstrup et~al.(2017)Svenstrup, Hansen, and
  Winther]{https://doi.org/10.48550/arxiv.1709.03933}
Svenstrup, D., Hansen, J.~M., and Winther, O.
\newblock Hash embeddings for efficient word representations, 2017.
\newblock URL \url{https://arxiv.org/abs/1709.03933}.

\bibitem[Taylor et~al.(2022)Taylor, Kardas, Cucurull, Scialom, Hartshorn,
  Saravia, Poulton, Kerkez, and
  Stojnic]{https://doi.org/10.48550/arxiv.2211.09085}
Taylor, R., Kardas, M., Cucurull, G., Scialom, T., Hartshorn, A., Saravia, E.,
  Poulton, A., Kerkez, V., and Stojnic, R.
\newblock Galactica: A large language model for science, 2022.
\newblock URL \url{https://arxiv.org/abs/2211.09085}.

\bibitem[Touvron et~al.(2023)Touvron, Lavril, Izacard, Martinet, Lachaux,
  Lacroix, Rozière, Goyal, Hambro, Azhar, Rodriguez, Joulin, Grave, and
  Lample]{llama}
Touvron, H., Lavril, T., Izacard, G., Martinet, X., Lachaux, M.-A., Lacroix,
  T., Rozière, B., Goyal, N., Hambro, E., Azhar, F., Rodriguez, A., Joulin,
  A., Grave, E., and Lample, G.
\newblock Llama: Open and efficient foundation language models, 2023.
\newblock URL \url{https://arxiv.org/abs/2302.13971}.

\bibitem[Wang et~al.(2022)Wang, Yuan, Rimanic, He, Dao, Chen, Re, and
  Zhang]{https://doi.org/10.48550/arxiv.2206.01299}
Wang, J., Yuan, B., Rimanic, L., He, Y., Dao, T., Chen, B., Re, C., and Zhang,
  C.
\newblock Fine-tuning language models over slow networks using activation
  compression with guarantees, 2022.
\newblock URL \url{https://arxiv.org/abs/2206.01299}.

\bibitem[Wortsman et~al.(2022)Wortsman, Gururangan, Li, Farhadi, Schmidt,
  Rabbat, and Morcos]{lofi}
Wortsman, M., Gururangan, S., Li, S., Farhadi, A., Schmidt, L., Rabbat, M., and
  Morcos, A.~S.
\newblock lo-fi: distributed fine-tuning without communication, 2022.
\newblock URL \url{https://arxiv.org/abs/2210.11948}.

\bibitem[Yang et~al.(2021)Yang, Hu, Babuschkin, Sidor, Liu, Farhi, Ryder,
  Pachocki, Chen, and Gao]{NEURIPS2021_8df7c2e3}
Yang, G., Hu, E., Babuschkin, I., Sidor, S., Liu, X., Farhi, D., Ryder, N.,
  Pachocki, J., Chen, W., and Gao, J.
\newblock Tuning large neural networks via zero-shot hyperparameter transfer.
\newblock In Ranzato, M., Beygelzimer, A., Dauphin, Y., Liang, P., and Vaughan,
  J.~W. (eds.), \emph{Advances in Neural Information Processing Systems},
  volume~34, pp.\  17084--17097. Curran Associates, Inc., 2021.
\newblock URL
  \url{https://proceedings.neurips.cc/paper/2021/file/8df7c2e3c3c3be098ef7b382bd2c37ba-Paper.pdf}.

\bibitem[Yuan et~al.(2022)Yuan, He, Davis, Zhang, Dao, Chen, Liang, Re, and
  Zhang]{https://doi.org/10.48550/arxiv.2206.01288}
Yuan, B., He, Y., Davis, J.~Q., Zhang, T., Dao, T., Chen, B., Liang, P., Re,
  C., and Zhang, C.
\newblock Decentralized training of foundation models in heterogeneous
  environments, 2022.
\newblock URL \url{https://arxiv.org/abs/2206.01288}.

\bibitem[Zhang et~al.(2022)Zhang, Roller, Goyal, Artetxe, Chen, Chen, Dewan,
  Diab, Li, Lin, Mihaylov, Ott, Shleifer, Shuster, Simig, Koura, Sridhar, Wang,
  and Zettlemoyer]{opt}
Zhang, S., Roller, S., Goyal, N., Artetxe, M., Chen, M., Chen, S., Dewan, C.,
  Diab, M., Li, X., Lin, X.~V., Mihaylov, T., Ott, M., Shleifer, S., Shuster,
  K., Simig, D., Koura, P.~S., Sridhar, A., Wang, T., and Zettlemoyer, L.
\newblock Opt: Open pre-trained transformer language models, 2022.
\newblock URL \url{https://arxiv.org/abs/2205.01068}.

\bibitem[Zhang et~al.(2016)Zhang, Zhao, and LeCun]{zhang2016characterlevel}
Zhang, X., Zhao, J., and LeCun, Y.
\newblock Character-level convolutional networks for text classification, 2016.

\bibitem[Zhou et~al.(2022)Zhou, Lei, Liu, Du, Huang, Zhao, Dai, Chen, Le, and
  Laudon]{https://doi.org/10.48550/arxiv.2202.09368}
Zhou, Y., Lei, T., Liu, H., Du, N., Huang, Y., Zhao, V., Dai, A., Chen, Z., Le,
  Q., and Laudon, J.
\newblock Mixture-of-experts with expert choice routing, 2022.
\newblock URL \url{https://arxiv.org/abs/2202.09368}.

\end{thebibliography}
\bibliographystyle{icml2023}


\newpage
\appendix
\onecolumn
\section{Appendix} \label{sec:appendix}
\begin{figure*}[h!]
\begin{subfigure}
  \centering
  \includegraphics[width=\textwidth]{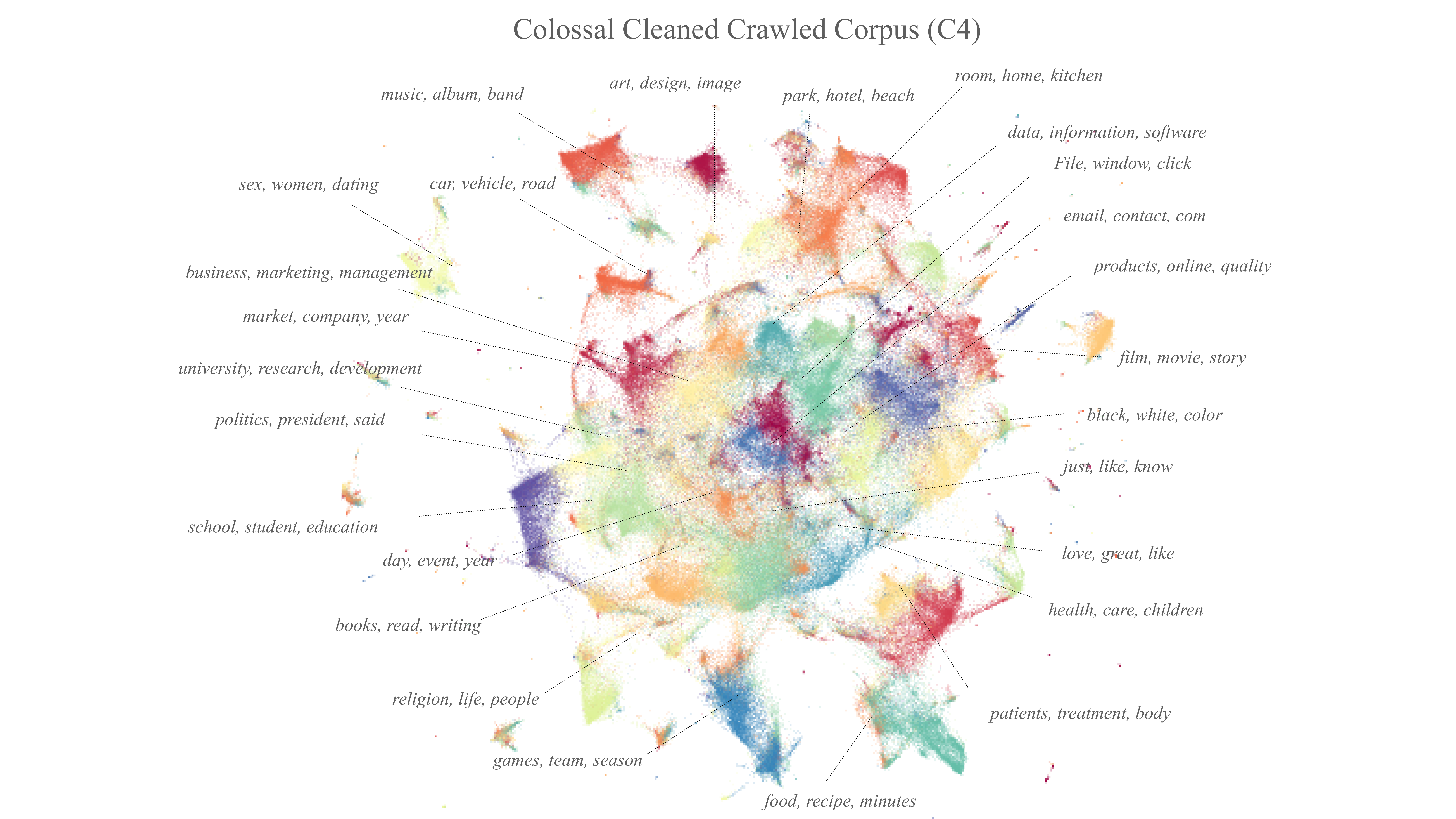}
  \label{fig:c4}
\end{subfigure}%
\begin{subfigure}
  \centering
  \includegraphics[width=\textwidth]{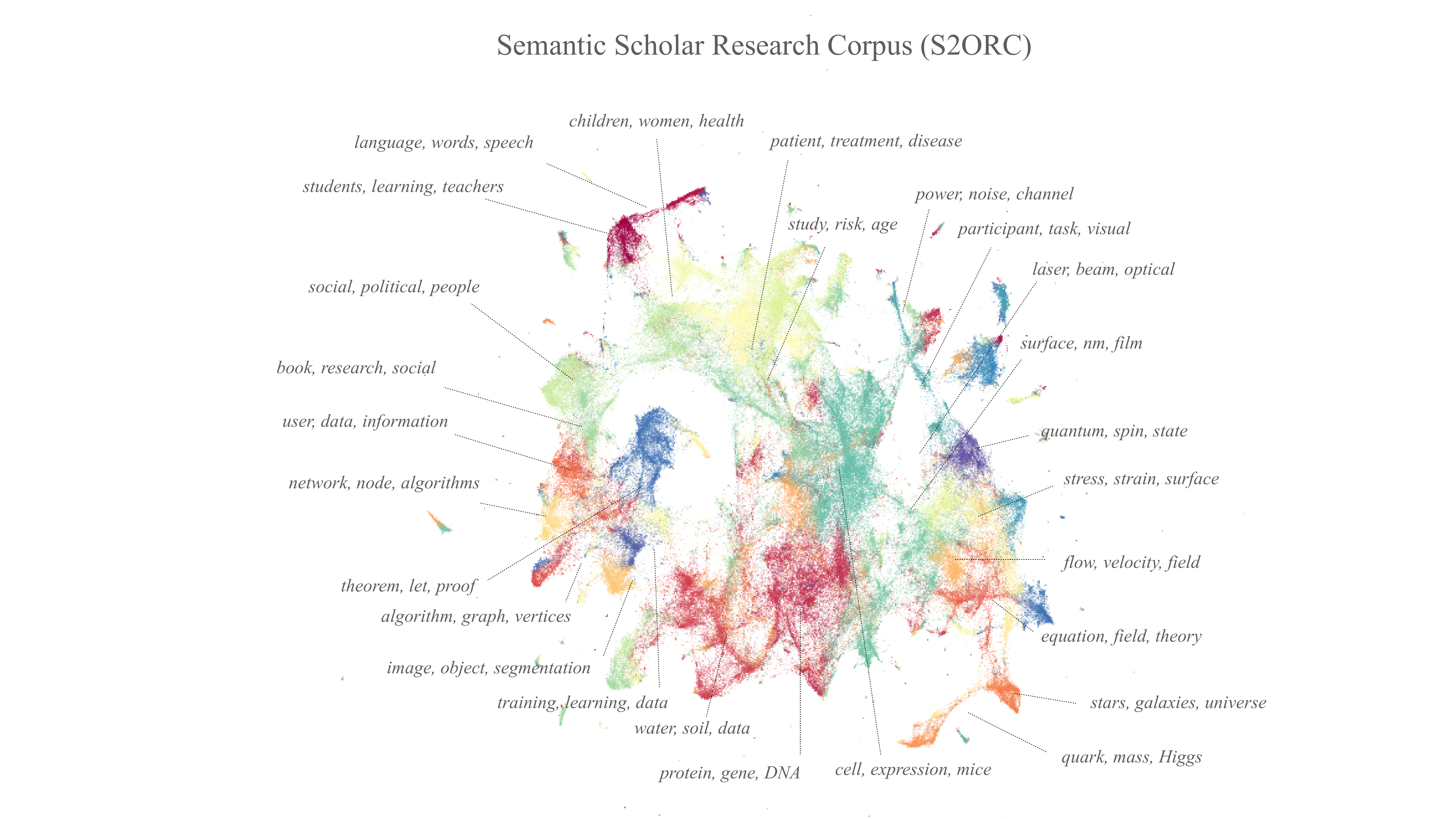}
  \label{fig:s2orc}
\end{subfigure}
\caption{Fully annotated UMAP visualization of 32 clusters in C4 and S2ORC. We annotate the clusters using an inverse transformation from our cluster centers back into the tf-idf vocabulary space, identifying the most likely words to generate the cluster center. We display the top 3 terms associated with each cluster center here.}
\label{fig:full_annotations}
\end{figure*}

\subsection{Clusters} \label{sec:cluster_terms}
\begin{table}[h!]
    \centering
    \begin{tabular}{cccccc}
\toprule
\bf Cluster & \bf Term 0 & \bf Term 1 & \bf Term 2 & \bf Term 3 & \bf Term 4 \\ \midrule
0 & fig & energy & al & et & field \\ 
1 & data & patients & social & time & al \\ 
\bottomrule
\end{tabular}
    \caption{Top terms associated with each cluster in S2ORC (2 clusters)}
    \label{tab:2cluster_s2orc}
\end{table}
\begin{table}[h!]
    \centering
    \begin{tabular}{cccccc}
\toprule
\bf Cluster & \bf Term 0 & \bf Term 1 & \bf Term 2 & \bf Term 3 & \bf Term 4 \\ \midrule
0 & eq & energy & quantum & equation & field \\ 
1 & algorithm & image & model & data & noise \\ 
2 & cells & cell & fig & al & protein \\ 
3 & students & social & people & research & education \\ 
4 & patients & patient & study & participants & children \\ 
5 & fig & temperature & energy & surface & beam \\
6 & user & data & node & network & nodes \\
7 & et & al & stars & galaxies & velocity \\
\bottomrule
\end{tabular}
    \caption{Top terms associated with each cluster in S2ORC (8 clusters)}
    \label{tab:8cluster_s2orc}
\end{table}
\begin{table*}[h!]
    \small
    \centering
    \begin{tabular}{cccccc}
\toprule
\bf Cluster & \bf Term 0 & \bf Term 1 & \bf Term 2 & \bf Term 3 & \bf Term 4 \\ \midrule
0 & students & learning & teachers & teaching & education \\ 
1 &language & word & words & speech & english \\ 
2 & network & node & nodes & networks & algorithm \\ 
3 & study & risk & age & data & group \\ 
4 & power & noise & channel & signal & frequency \\ 
5 & quantum & spin & state & magnetic & states \\ 
6 & participants & task & visual & stimulus & et  \\ 
7 & theorem & let & proof & lemma & set \\ 
8 & training & learning & data & network & model \\ 
9 & laser & beam & optical & fig & pulse \\ 
10 & protein & genes & gene & dna & proteins \\ 
11 & water & soil & data & et & al \\ 
12 & algorithm & graph & vertices & problem & vertex \\
13 & flow & velocity & field & magnetic & wave \\
14 & user & data & users & information & service \\
15 & stars & galaxies & et & al & star \\
16 & quark & mass & gev & higgs & energy \\
17 & et & al & brain & neurons & fig \\
18 & stress & strain & surface & shear & fig \\
19 & image & images & algorithm & object & segmentation \\
20 & energy & electron & fig & eq & state \\
21 & equation & field & eq & equations & theory \\
22 & patients & patient & treatment & study & disease \\
23 & model & time & robot & state & control \\
24 & children & child & women & health & social \\
25 & surface & nm & film & layer & graphene \\
26 & patient & patients & pain & surgery & treatment \\
27 & social & political & people & policy & public \\
28 & social & participants & health & self & people \\
29 & book & research & social & information & data \\
30 & temperature & water & heat & phase & thermal \\
31 & cells & cell & expression & mice & fig \\
\bottomrule
\end{tabular}
    \caption{Top terms associated with each cluster in S2ORC (32 Clusters).}
    \label{tab:32cluster_s2orc}
\end{table*}

\begin{table}[!ht]
    \centering
    \begin{tabular}{cccccc}
\toprule
\bf Cluster & \bf Term 0 & \bf Term 1 & \bf Term 2 & \bf Term 3 & \bf Term 4 \\ \midrule
0 & like & just & time & new & great \\ 
1 & new & information & use & business & services \\ 
\bottomrule
\end{tabular}
    \caption{Top terms associated with each cluster in C4 (2 clusters)}
    \label{tab:2cluster_c4}
\end{table}
\begin{table}[h!]
    \centering
    \begin{tabular}{cccccc}
\toprule
\bf Cluster & \bf Term 0 & \bf Term 1 & \bf Term 2 & \bf Term 3 & \bf Term 4 \\ \midrule
0 & students & health & research & school & university \\
1 & music & new & love & film & art \\
2 & said & year & state & team & new \\
3 & business & company & services & service & market \\
4 & like & just & time & don & really \\
5 & use & information & data & page & click \\
6 & design & black & white & color & high \\
7 & home & water & food & park & area \\ 
\bottomrule
\end{tabular}
    \caption{Top terms associated with each cluster in C4 (8 clusters)}
    \label{tab:8cluster_c4}
\end{table}
\begin{table*}[h!]
    \small
    \centering
    \begin{tabular}{cccccc}
\toprule
\bf Cluster & \bf Term 0 & \bf Term 1 & \bf Term 2 & \bf Term 3 & \bf Term 4 \\ \midrule
0 & site & page & website & web & search \\ 
1 & art & design & image & images & gallery \\ 
2 & health & care & children & child & medical \\ 
3 & just & like & don & know & ve \\ 
4 & data & information & software & use & management \\ 
5 & love & great & like & just & really \\ 
6 & game & team & games & season & play \\ 
7 & information & email & contact & com & address \\ 
8 & power & high & use & light & steel \\ 
9 & students & school & student & education & learning \\ 
10 & market & company & year & financial & tax \\ 
11 & patients & treatment & body & cancer & pain \\ 
12 & room & home & kitchen & bedroom & living \\ 
13 & music & album & band & song & songs \\ 
14 & car & vehicle & cars & new & road \\ 
15 & park & hotel & beach & area & travel \\ 
16 & day & event & year & time & wedding \\ 
17 & service & services & quality & customer & best \\ 
18 & book & books & read & writing & story \\ 
19 & film & movie & story & new & like \\ 
20 & black & white & color & dress & look \\ 
21 & business & company & marketing & management & customers \\ 
22 & university & research & development & education & science \\ 
23 & city & community & county & said & police \\ 
24 & sex & women & porn & dating & girls \\ 
25 & products & product & online & quality & order \\ 
26 & religion & life & people & jesus & time \\ 
27 & water & skin & use & oil & like \\ 
28 & said & politics & president & government & state \\ 
29 & app & phone & video & mobile & casino \\ 
30 & file & windows & use & click & software \\ 
31 & food & add & recipe & minutes & wine \\ 
\bottomrule
\end{tabular}
    \caption{Top terms associated with each cluster in C4 (32 Clusters).}
    \label{tab:32cluster_c4}
\end{table*}

\subsection{Comparing FLOP counts via training data size}
\label{sec:comparing_flop_counts}

To make fair comparisons across models with different numbers of ELMs, for a given number of clusters $k$ and total GPU budget $n$, each ELM is allocated $n/k$ GPUs. This keeps the total effective number of FLOPs fixed across models exposed to the same number of tokens. We can show this analytically; following \citealt{artetxe2021efficient}, we calculate the number of FLOPs to train a single ELM in our experiments:

\[ F_{\text{ELM}(T, k)} = \frac{96lh^2T}{k} \left (1 + \frac{s}{6h}  + \frac{V}{16lh} \right) \]

where $T$ is the total training tokens (i.e., sequence length $\times$ batch size per GPU $\times$ number of GPUs), $k$ is the number of clusters, $l$ is the number of layers, $h$ is the hidden dimension, $s$ is the sequence length, and $V$ is the vocabulary.

Therefore, the total cost in FLOPs to train $k$ ELMs with a particular model architecture (e.g., OPT-1.3B) on $T$ tokens in aggregate is equivalent to that of a single dense LM of the same architecture trained on $T$ tokens:

\[ k \cdot F_{\text{ELM}(T, k)} = F_{\text{ELM(T, 1)}} \]

This means that even though \cbtm trains $k$ times more parameters than an equivalent dense model, it does so \emph{at the same overall cost in FLOPs}. So, our comparisons of models with various numbers of ELMs are fair, as long as they have been exposed to the same number of training tokens and have the same underlying architecture for each ELM. Therefore, throughout the paper, we use training data size as a more interpretable metric of the overall compute budget.

\subsection{Interpolating between empirical observations when comparing training costs and performance}
\label{sec:perf_interpolation}

We interpolate between our empirical observations using the following function, proposed in \citealt{artetxe2021efficient}:

\[ c(t)  = \exp (\log c_{lo}(t) + r (
\log c_{hi}(t) - \log c_{lo}(t)))\]

where $r=\frac{t - t_{lo}}{t_{hi} - t_{lo}}$, $t_{hi}$ and $t_{lo}$ are the closest empirical performances to $t$ and $c_{lo}(t)$ and $c_{hi}(t)$ are their corresponding training cost in FLOPs. We use this interpolation to compute the speedup factor $c_{dense}(t) / c_{cbtm}(t)$.

\subsection{Effect of cluster balancing on cluster sizes}
\label{sec:appendix_balancing}

\begin{figure}[h!]
    \centering
    \includegraphics[scale=0.5]{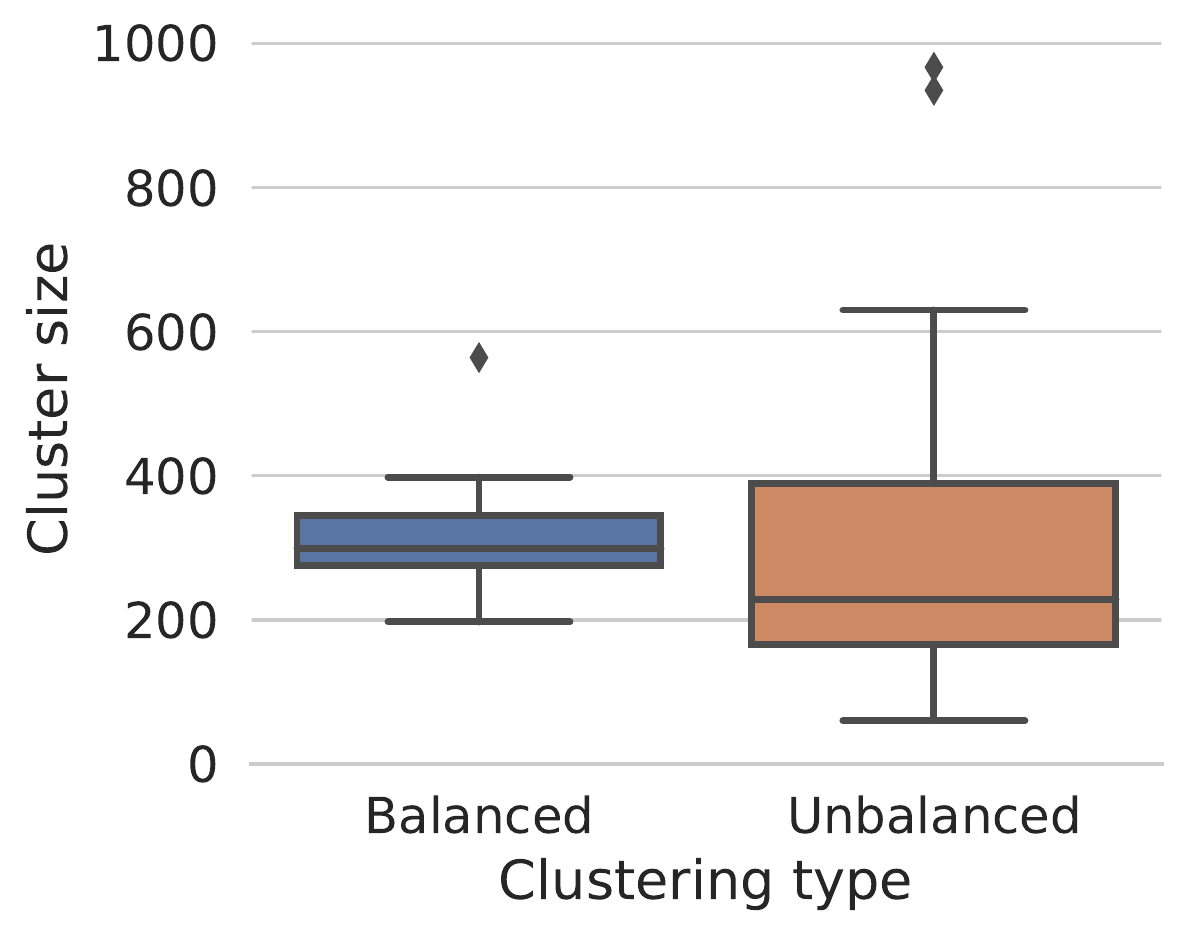} \vspace{-1em}
    \caption{\textbf{Cluster balancing narrows the range, and increases the median size, of clusters  (\S\ref{sec:analysis_balancing}).} Here, we ablate our balancing procedure from \S\ref{sec:cbtm_iteration} on a 10K held-out documents in C4.}
    \label{fig:cluster_sizes}
\end{figure}

Using a held out set of 10K documents from C4, we ablate our balancing procedure from \S\ref{sec:cbtm_iteration}, and display a boxplot showing cluster sizes in Figure \ref{fig:cluster_sizes}.

\subsection{The Pile experiments}
\label{sec:metadata_comparison}

For the experiment in \S\ref{sec:analysis_metadata}, we additionally train on The Pile, which is a publicly available corpus of diverse language modeling datasets. We use the filtered version from the OPT pretraining data \citep{opt}, and subselect 8 datasets of the 13 used for OPT pretraining, which enabled easier experimentation \S\ref{sec:analysis_metadata}. These 8 datasets include: Common Crawl, HackerNews comments, Reddit comments\footnote{Reddit comments, like the rest of these datasets, are not collected by us but were part of the Pile \citep{pile}, a publicly available third-party dataset.}, the Gutenberg corpus, STORIES corpus, OpenWebText2, Deepmind-Mathematics, and English Wikipedia. In aggregate, these datasets consist of 420M documents, totaling 116B BPE tokens. For evaluation, we sample an equal number of documents from each constituent dataset. We also train a $k$=8 clustering model on this dataset with one shard, or 1.5M documents.

\begin{table}[h!]
\centering
\small
\begin{tabular}{ccc}
\multicolumn{3}{c}{\textbf{5.2B Tokens}} \\
\toprule
  \bf 8 Metadata  & \bf 8 Clusters &  \bf 1 Cluster  \\
\midrule 
8.4 & 8.3 & 8.5 \\
\bottomrule
\end{tabular}

\caption{\textbf{Experts trained with clusters perform slightly better than experts trained with metadata (\S\ref{sec:analysis_metadata}).} Here, we train each model for 5.2B tokens on 8 corpora of the Pile, and evaluate perplexity on a held out random sample of the constituent datasets. See \S\ref{sec:metadata_comparison} for more details on the dataset.}
\label{tab:metadatacompare}
\end{table}

\subsection{Sparsity}
\label{sec:appendix_sparsity}

\begin{table}[h!]
\centering
\small
\begin{tabular}{cc|c|cc|cccc|cccccc}
\toprule
& & \bf Dense   & \multicolumn{2}{c|}{\bf 2-cluster} & \multicolumn{4}{c|}{\bf 8-cluster} & \multicolumn{6}{c|}{\bf 32-cluster} \\
& & &  top-1 & top-2 &   top-1 & top-2 & top-4 & top-8 &   top-1 & top-2 & top-4 & top-8 & top-16 & top-32 \\
\cmidrule{1-15}
\parbox[t]{3pt}{\multirow{7}{*}{\rotatebox[origin=c]{90}{\bf {\textsc{Temperature}}}}} 
& 0.01	& \bf 13.82	& \bf 13.64	& 13.60	& \bf 13.64	& 13.50	& 13.49	& 13.49	& \bf 13.55	& 13.45	& 13.45	& 13.45	& 13.45	& 13.45 \\
& 0.05	& -	& -	    & 13.51	& -	& 13.32	& 13.25	& 13.25	& -	& 13.25	& 13.19	& 13.17	& 13.17	& 13.17 \\
& 0.1	    & -	& -      & \bf 13.50	& -	& \bf 13.27	& \bf 13.22	& \bf 13.24   & -	& 13.16	& \bf 13.05	& \bf 13.00	& \bf 13.01	& \bf 13.01\\
& 0.2	    & -	& -      & 13.54	& -	& 13.29	& 13.34	& 13.50	& -	& \bf 13.14	& 13.06	& 13.07	& 13.07	& 13.28 \\
& 0.3	    & - & -  	& 13.59	& -	& 13.33	& 13.45	& 13.72	& -	& 13.17	& 13.15	& 13.27	& 13.54	& 13.78 \\
& 0.5	    & - & -	    & 13.64	& -	& 13.38	& 13.59	& 13.97	& -	& 13.22	& 13.30 	& 13.57	& 14.02	& 14.46 \\
& 1	    & - & -    &	13.69	& -	& 13.43	& 13.73	& 14.19	& -	& 13.30	& 13.49	& 13.89	& 14.46	& 15.04 \\

\bottomrule

\end{tabular}

\caption{Results with the dense (1-cluster), 2-cluster, 8-cluster, and 32-cluster C4 models trained on 84B tokens when varying the temperature ($T$) and $topk$ hyperparameters. Optimal performance for almost every model and $topk$ value is achieved at $T=0.1$.
}
\label{tab:inference}
\end{table}

\subsection{Downstream tasks}
\label{sec:appendix_downstream_tasks}

\paragraph{Tasks} We experiment with six text classification tasks which span topic classification (AGNews; \citealt{zhang2016characterlevel} and DBPedia; \citealt{dbpedia}); sentiment analysis (SST-2; \citealt{maas-etal-2011-learning}, Amazon; \citealt{zhang2016characterlevel}, and Phrasebank \citealt{Malo2014GoodDO}); and hatespeech detection (Twitter; \citealt{barbieri-etal-2020-tweeteval}).

\paragraph{Performance Routing}

We introduce additional routing procedures for few-shot text classification, as the clustering method described in \S\ref{sec:cbtm_inference} ignores the significance of labels and does not take into account the context's word order, which may be crucial when the context contains demonstrations for a downstream task. Further, the specific ordering of in-context demonstrations is known to affect model performance~\cite{lu-etal-2022-fantastically}. There is not sufficient evidence to suggest that the relative rank of different models on a task stays constant through these performance variances; thus it may be important to route to experts differently depending on demonstration example order. To take into account the order of tokens in the context, we introduce 3 variations on routing, based on expert performance on the end task, which take inspiration from the mixture probability estimation methods of \citet{gururangan-etal-2022-demix,btm}. 

To perform \textit{Fixed Performance Routing with demonstrations and validation set examples}, we select 8 examples randomly from the validation set, such that there is no overlap with the 8 demonstration examples used in the context at test time. We concatenate the 8 demonstrations with one validation set context and evaluate the accuracy of each \elm on the sequence, repeating for each of the 8 examples from the validation set. The final routing probability distribution over experts for this task is determined by a softmax over the average accuracy of the model over the 8 examples. We use this fixed probability distribution for all test examples. 

To perform \textit{Fixed Performance Routing with only demonstrations}, we adapt the procedure above such that no validation examples are necessary. Instead, we take a random permutation of the 8 demonstration examples. We remove the label of the last, such that we effectively use the first 7 as demonstrations, and rely on the last example to estimate performance. We repeat this 8 times, generating a new random permutation each time, and once again take the softmax over the average accuracy of the model for each permutation, fixing this distribution for all test examples.

Finally, we have \textit{Updating Performance Routing}, in which no estimations are done before test-time. At test time, we begin with a uniform probability over all experts, which we update with an exponential moving average with each test example: after each example (prepended with the 8 demonstrations), we update the expert probabilities with the softmax over the accuracy of each expert on that example. Once again, this distribution is fixed for all test examples.

\paragraph{Results} Full results, in Table~\ref{tab:downstream_tasks_appendix}, show that \textit{Fixed Performance Routing with demonstrations and validation set examples} achieves the best performance overall, with optimal performance occurring at top-4 expert activation, which we also found in the language modeling results of \S\ref{sec:core_results}. Both \textit{Fixed Performance Routing} methods perform best with top-4 expert activation, and only suffer small performance degradations when reduced to top-1 expert activation. This aligns well with the patterns observed in \S\ref{sec:core_results}, which further supports the incorporation of end task performance in routing when adapting to new tasks, even when we base this evaluation only on the demonstration examples -- that is, without any additional data. We leave for future work further tuning of the optimal settings for Performance Routing.




\begin{table*}[t]
\centering
\small
\begin{tabular}{rrrr}
\toprule
\textbf{Task}  & \textbf{1st} & \textbf{2nd} & \textbf{3rd} \\
\midrule
AGNews & \emph{game, team, season} & \emph{said, government, president} & \emph{business, company, market}  \\
DBPedia & \emph{students, school, university} & \emph{said, government, president} & \emph{city, park, hotel} \\
SST-2 & \emph{book, film, life} & \emph{love, family, great} & \emph{music, art, band} \\
Amazon & \emph{book, film, life} & \emph{just, like, know} & \emph{game, team, season} \\
Phrasebank & \emph{business, company, market} & \emph{service, customer, products} & \emph{said, government, president} \\
Twitter & \emph{data, software, download} & \emph{click, website, page} &  \emph{just, like, know} \\







\bottomrule \\
\end{tabular}

\caption{\textbf{Top-terms of clusters associated with top-3 experts for each classification task (\S\ref{sec:downstream_task_results})} By inspection, the highest probability experts are usually quite relevant to task's domain.}
\label{tab:topk_examples}
\end{table*}

\begin{table*}[t!]
\centering
\small
\begin{tabular}{lrrrrrrr}
& \multicolumn{6}{c}{\it Few-shot Text Classification Accuracy (\%)} \\
\toprule
  & \bf AGNews & \bf DBPedia & \bf SST-2 & \bf Amazon & \bf Phrasebank  & \bf Twitter \\ 
 $\downarrow$ Model (inference parameters) &  \it Topic & \it Topic & \it Sentiment & \it Sentiment & \it Sentiment & \it Hatespeech & \bf Average \\

\midrule
Random chance & 25.0 & 7.10 & 50.0 & 50.0 & 33.3 & 50.0 & 35.9 \\
OPT (1.3B) &  42.9 & 57.2 & 72.8 & 81.3 & 72.5 & 65.1 & 65.3 \\

OPT (6.7B) &  51.9 & 58.9 & 77.0 & 83.8 & 76.4 & 39.6 & 64.6 \\
\midrule
1-cluster (1.3B) & 47.4 & 61.1 & 80.2 & 80.7 & 66.6 & 60.9 & 66.2 \\
1-cluster (6.7B) & \bf 68.1 & 62.4 & 80.7 & \bf 84.9 &  80.6 & 37.4 & 69.0 \\
\midrule

 & \multicolumn{6}{c}{\textit{Cluster Routing}}
  \vspace{1mm}
\\
16-cluster; top-1 (1.3B) & 47.1	& 62.9	& 74.3 & 	79.1 &	72.9 &	56.4 & 65.4  \\
16-cluster; top-4 (5.2B) & 49.3	& 62.3 &  80.0 &	81.3 &	  78.7 & 	61.3 & 68.8 \\
16-cluster; top-16 (20.8B) & 50.6 &	62.0 &	84.0 & 83.2 &	78.6 &	 61.7 & 69.9  \\
\midrule
 & \multicolumn{6}{c}{\textit{Updating Performance Routing}}
  \vspace{1mm}
\\

16-cluster; top-1 (1.3B) & 54.5 &	\bf 63.4	& 83.4	&	83.6		&74.7		&64.8 & 63.3 \\
16-cluster; top-4 (5.2B) &51.6		&61.5		&88.6		&83.7		&\bf 80.7	& \bf 65.3	& 68.8	 \\
16-cluster; top-16 (20.8B) & 51.1		& 60.4	&	86.0	&	83.5	&	79.8	&	62.9	& 69.1	 \\
\midrule
 & \multicolumn{6}{c}{\textit{Fixed Performance Routing (8 demonstrations)}}
  \vspace{1mm}
\\
16-cluster; top-1 (1.3B) & 45.3	& 61.9 & 	81.2 & 	83.6 &	76.4 &	60.1 & 68.1 \\
16-cluster; top-4 (5.2B) & 51.2	& 60.9	 & 81.4 &	83.0 & 	80.5& 	60.9 & 69.6 \\
16-cluster; top-16 (20.8B) & 50.6	& 60.2 &	84.1 &	83.5	& 79.1	& 60.5 & 69.6 \\
\midrule
 & \multicolumn{6}{c}{\textit{Fixed Performance Routing (8 demonstrations + 8 validation examples)}}
  \vspace{1mm} \\
16-cluster; top-1 (1.3B) & 54.5	& \bf 63.4	& 83.4	 & 83.6	& 74.7 & 64.8 & 70.7 \\
16-cluster; top-4 (5.2B) & 51.6 &	61.5 &  \bf 	88.6	 & 83.7 &	\bf 80.7 &	\bf 65.3 & \bf 71.9 \\
16-cluster; top-16 (20.8B) & 51.1 &	60.4	& 86.0	& 83.5	& 79.8 &	 62.9 & 70.6\\ 

\bottomrule
\end{tabular}
\caption{\textbf{\cbtm models with performance routing achieve even better performance on downstream tasks (\S\ref{sec:appendix_downstream_tasks}).} We display performance of models on six text classification tasks, using eight demonstrations for each example and no additional fine-tuning. We compare our cluster routing method (described in \S\ref{sec:downstream_tasks}) to variants of performance routing (described in \S\ref{sec:appendix_downstream_tasks}). Fixed performance routing with 8 demonstrations and 8 validation examples usually gets the best performance on downstream tasks, consistently outperforming even the 6.7B parameter baselines. Fixed performance routing with top-4 inference always improves performance over using all experts, and top-1 inference  does substantially better than the dense baselines at no additional inference costs. We include average performance across tasks for readability.}
\label{tab:downstream_tasks_appendix}
\end{table*}

\end{document}